\documentclass[10pt,twocolumn,letterpaper]{article}

\usepackage{iccv}
\usepackage{times}
\usepackage{epsfig}
\usepackage{graphicx}
\usepackage{amsmath}
\usepackage{amssymb}
\usepackage{subfig}
\usepackage{float}
\usepackage{caption}
\usepackage{array}

\usepackage[pagebackref=true,breaklinks=true,letterpaper=true,colorlinks,bookmarks=false]{hyperref}

\newcommand{\bftab}{\fontseries{b}\selectfont}

\iccvfinalcopy % *** Uncomment this line for the final submission

\ificcvfinal\fi

\begin{document}

%%%%%%%%% TITLE
\title{SceneFormer: Indoor Scene Generation with Transformers}

\author{Xinpeng Wang$^{\ast}$\quad
	Chandan Yeshwanth$^{*}$\quad
	Matthias Nie{\ss}ner \quad\\
Technical University of Munich\\

}

\twocolumn[{%
\renewcommand\twocolumn[1][]{#1}%
\maketitle
\begin{center}

\centering
\vspace{-0.1cm}
\includegraphics[width=0.95\textwidth]{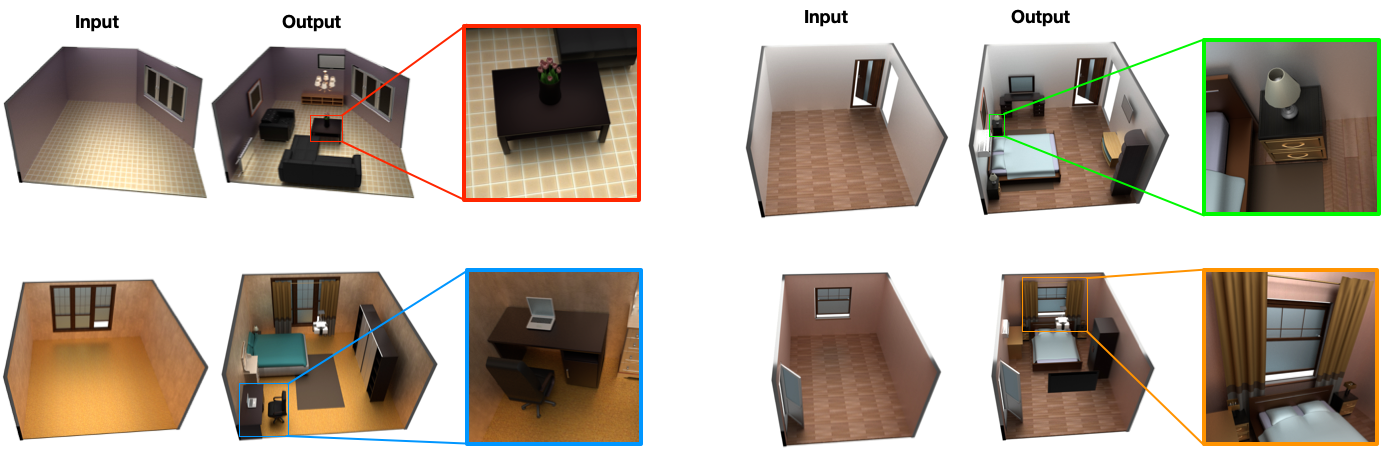}
\vspace{-0.2cm}
\captionof{figure}{3D scenes generated by our method: the input to our model is a room layout (e.g., room floor, locations of doors and windows), from which we populate the room with CAD objects to generate a full 3D scene. We predict a sequence of object locations, leveraging the self-attention mechanism of transformers to obtain realistic 3D scene arrangements.}

\end{center}
}]

\let\thefootnote\relax\footnotetext{$^{\ast}$ Equal contribution}

\ificcvfinal\thispagestyle{empty}\fi

%%%%%%%%% ABSTRACT
\begin{abstract}
   We address the task of indoor scene generation by  generating a sequence of objects, along with their locations and orientations conditioned on a room layout. Large-scale indoor scene datasets allow us to extract patterns from user-designed indoor scenes, and generate new scenes based on these patterns. Existing methods rely on the 2D or 3D appearance of these scenes in addition to object positions, and make assumptions about the possible relations between objects. In contrast, we do not use any appearance information, and implicitly learn object relations using the self-attention mechanism of transformers. We show that our model design leads to faster scene generation with similar or improved levels of realism compared to previous methods. Our method is also flexible, as it can be conditioned not only on the room layout but also on text descriptions of the room, using only the cross-attention mechanism of transformers. Our user study shows that our generated scenes are preferred to the state-of-the-art FastSynth scenes 53.9\% and  56.7\%  of the time for bedroom and living room scenes, respectively. At the same time, we generate a scene in $1.48$ seconds on average, $20\%$ faster than  FastSynth.
\end{abstract}

%%%%%%%%% BODY TEXT
\section{Introduction}
Generating realistic 3D indoor scenes has a wide range of real-world applications for 3D content creation. 
For instance, real estate and interior furnishing companies can quickly visualize a furnished room and its contents without requiring the rearrangement of any physical objects. 
Such a room can be presented through augmented or virtual reality platforms such as a headset, allowing a user to walk through their future home and interactively modify it.

We address the task of scene generation from a room layout by generating a set of objects and their arrangements in the room.
Each object is generated with a predicted class category, a 3D location, its angular orientation, and a 3D size. 
Once this sequence is generated, the most relevant CAD model for each object is retrieved from a database and placed in the scene at the predicted location. 
The relevance of a CAD model can be predicted based only on size, a shape descriptor  \cite{zhang2020deep}, or other heuristics such as texture.
CAD model selection reduces object collisions, accommodates special object properties such as symmetry and can ensure style-consistency across objects.

Existing methods operate on internal representations of the scene such as 2D images  \cite{wang2018deep}, graphs  \cite{wang2019planit}, and matrices  \cite{zhang2020deep}.
Several of these works generate objects in an autoregressive manner - the properties of the $(n+1)^{th}$ object are conditioned on the properties of the first $n$ objects. 
We adopt a similar autoregressive prediction. 
The extraction of object and scene patterns is enabled by large object and scene datasets such as ModelNet ~\cite{wu20153d}, ShapeNet~\cite{chang2015shapenet} and other human-created scene datasets with synthetic objects  \cite{song2017semantic}. 
Some existing methods require object relations to be annotated, and assume a fixed  set of possible relations  \cite{zhou2019scenegraphnet, wang2019planit}, instead of operating on the raw scene data. In contrast, we operate directly on the raw locations and orientations of the objects without any additional information. We thus avoid any bias introduced by the manual selection of relations, or the heuristics used to create these relations.

Transformers  \cite{vaswani2017attention} perform well on a variety of natural language processing tasks by treating a sentence as a sequence of words, and recently on images  \cite{luo2020end} and 3D meshes  \cite{nash2020polygen} as well. Based on the idea that a scene can be treated as a \textit{sequence} of objects, we propose \textit{SceneFormer}, a series of transformers that autoregressively predict the class category, location, orientation and the size of each object in a scene.  We show that such a model generates realistic and diverse scenes, while requiring little domain knowledge or data preparation. In addition, we do not use any visual information such as a 2D rendering of the scene either as input or as an internal representation.

We use the cross-attention mechanism of the Transformer decoder to build conditional models. Conditioning differs from a translation task -- in translation the output is expected to be aligned with the input as closely as possible according to a given metric, while in a conditioning task the  input only guides the scene generation. We expect the output to mostly adhere to the input, but it can contain some information not indicated by the input.
We condition separately on two kinds of user inputs. The first is the room layout, including the positions of doors and windows. Our layout-conditioned scenes are preferred over the state-of-the-art FastSynth 53.9\% of the time for bedroom scenes, and 56.7\% for living room scenes. The second type of input is a text description of the room, such as \textit{``There is a room with a bed and a wardrobe. There is a table lamp next to the wardrobe.''}. Our text-conditioned model outperforms baselines in terms of category and relation accuracy.

In summary, our main contributions are:
\begin{itemize}

\item We represent an indoor scene as a \textit{sequence} of object properties, converting scene generation to a sequence generation task.

\item We leverage the self-attention of transformers to implicitly learn relations between objects in a scene, eliminating the need for manually-annotated relations.

\item We generate complex scenes conditioned on room layout or text descriptions by leveraging discretized object coordinates to predict their 3D locations.

\end{itemize}

\begin{figure*}[h!]
\vspace{-0.1cm}
\centering
\includegraphics[width=0.95\textwidth]{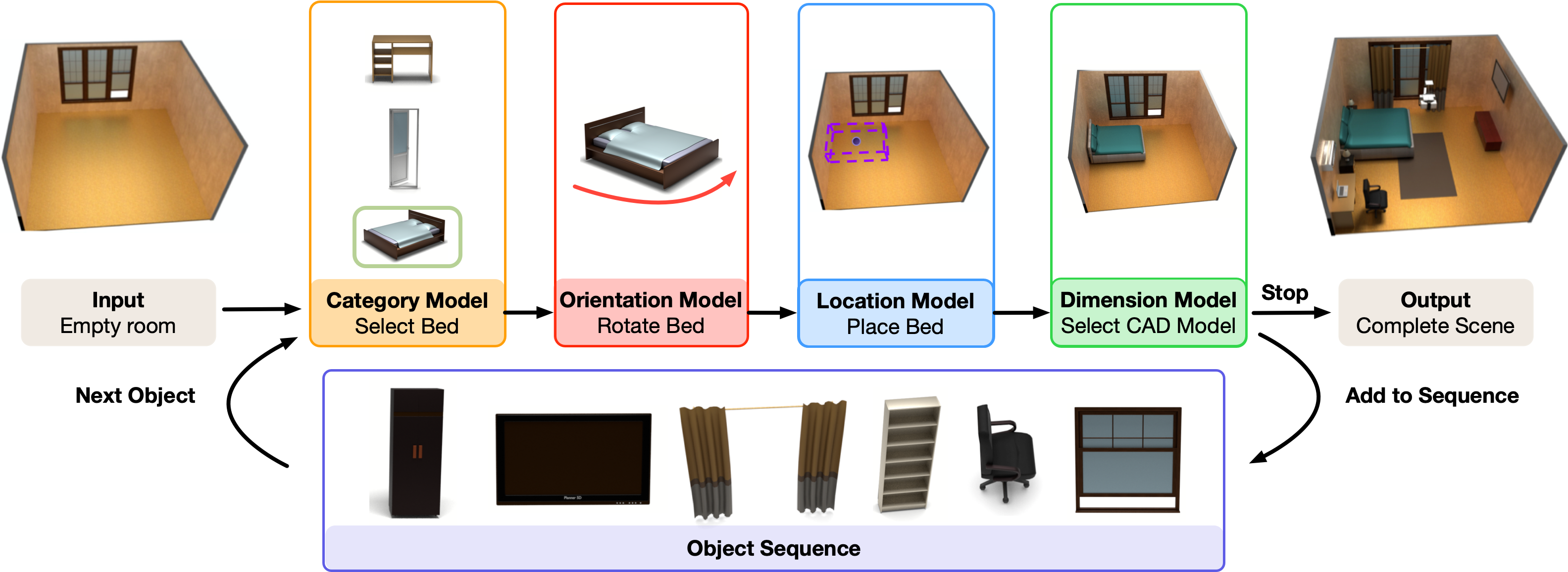}
\vspace{-0.1cm}
\caption{Layout-conditioned SceneFormer: We take as input the room layout describing the room shape and locations of doors and windows. The SceneFormer model sequentially generates the properties of the next object and inserts the object into the existing scene. The final output scene is shown on  the right. }
\label{fig:pipeline}
\end{figure*}

\section{Related Work}

\paragraph{Graph-based 3D Scene Generation.}
A natural representation of a scene is a graph   \cite{zhou2019scenegraphnet, luo2020end, purkait2019learning}, where each object is a node, and an edge is a relation between objects (e.g., `is next to`, `is on top of`). Features of the room such as walls, doors and windows can be represented as nodes of the graph  \cite{wang2019planit}. This gives a simple method for conditioning -- when the model is autoregressive on the graph and is able to generate one node at a time, the input is initialized with the required object nodes and then repeatedly expanded using the model. Such a representation lends itself well to processing with graph convolutional networks \cite{kipf2017semi}.

A scene can also be represented as a matrix  of objects in each category \cite{zhang2020deep} or a hierarchy  of objects \cite{li2019grains}. Next, an appropriate encoder-decoder or other model is used to generate an output scene in the same format. Other methods place objects with a Bayesian model \cite{liang2017automatic}, or model object properties with probability distributions learned from the data \cite{henderson2017automatic, merrell2011interactive}. Yet other works condition on full 3D scans \cite{kermani2016learning}, or focus on fine-grained and smaller objects such as a table with multiple objects on it \cite{fisher2012example}. However, these methods require complex optimization and post-processing steps to obtain realistic scenes. This may be in the form of a discriminative loss, linear programming or other heuristics that are specific to the task. In contrast, we use only the cross-entropy loss for classification that is both conceptually simpler and easier to optimize.

\paragraph{Scene Generation from an Image.}
A scene can be represented by a top down view of the objects, walls, and floor  \cite{ritchie2019fast, wang2018deep}.  \cite{ritchie2019fast} and \cite{wang2018deep} predict the walls and floors as binary images, and object properties by their continuous or discrete values. 
This can be used to represent arbitrary room shapes and sizes, by normalizing the room dimensions to a known maximum dimension along each axis. Image representations can take advantage of modern CNN architectures such as ResNet~\cite{he2016deep}.

\paragraph{Text-Conditioned Scene Generation.}
Several previous works have addressed the task of text-conditional generation or text-to-scene translation. 
Text inputs have been used to create detailed partial scenes \cite{ma2018language, chang2014semantic}, such as a table with several small objects on it, which are then inserted into the larger scene. User input is required for refinement at every step, making it semi-automatic. Similarly, SceneSeer \cite{chang2017sceneseer} and related methods \cite{chang2014interactive, savva2017scenesuggest} rely on interactive user inputs. 
Other related works  \cite{chang2015text, chang2014learning} generate simple scenes with few objects in them, by inferring rules from a smaller-scale human-annotated dataset. Text2Scene \cite{tan2019text2scene} solves a similar task in 2D, by iteratively placing objects into an 2D image and then ensuring consistency. 
Intelligent Home 3D~\cite{chen2020intelligent} tackles the related task of generating the full room layout of a house from text, and proposes a new dataset for this task.
Our method differs from these in that it can generate high quality complex scenes with a large number of objects, without requiring user input. However, our model is still flexible enough to accept user inputs if desired.

\begin{figure*}
\vspace{-0.1cm}
\centering
\includegraphics[width=0.85\textwidth]{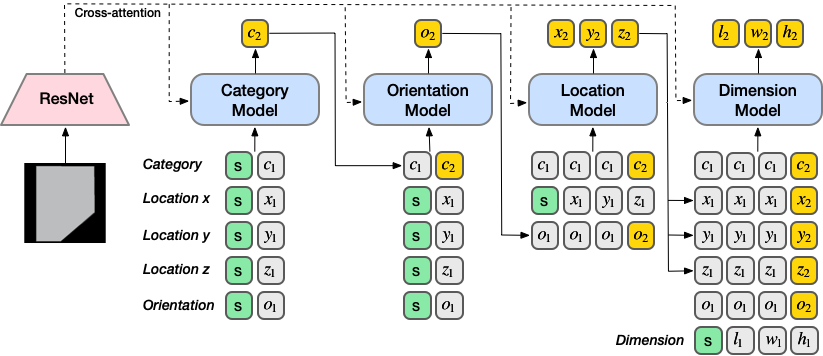}
\vspace{-0.2cm}
\caption{Layout-conditioned SceneFormer model. Start tokens are shown in green, tokens for existing objects in gray and tokens for new objects in yellow. Stop tokens and padding are omitted. All models take 3 kinds of sequences as input - category, orientation and location. Their respective outputs, except in the case of the dimension model, are appended to the existing sequence before passing it on. A model with $N$ output tokens is run $N$ times, producing one token from each step.}
\label{fig:sceneformer}
\end{figure*}

\section{Method}
An overview of our scene generation approach is shown in Fig.~\ref{fig:pipeline}. We first discuss our data preparation, and then the \textit{SceneFormer} model and its layout-conditioned and text-conditioned variants.

\subsection{Data Preparation}
We treat each scene as a sequence of objects, ordered by the frequency $f_{c_i}$ of their class categories $c_i$ in the train set. 
This ordering of objects is required to produce a unique representation of each scene, up to the ordering within objects of the same class. 
Then the location of an object is normalized by the maximum room size, and quantized into the range $[0, 255]$ to give the new coordinates of the object $(x, y, z)$. 
Similarly the dimensions of each object; length, width and height or $(l, w, h)$ are scaled and quantized. The orientation of the object in the floor plane of the room $\theta$ is quantized in the range $[0, 359]$. 
Hence, for each scene with objects $\{o_i\}$ we obtain 8 sequences $(\{c_i\}, \{x_i\}, \{y_i\}, \{z_i\}, \{\theta_i\}, \{l_i\}, \{w_i\}, \{h_i\})$. We then add  $start$ and $stop$ tokens to each sequence, indicating the beginning and end, and finally pad the sequence to the maximum length present in the train set.

\subsection{Transformer for Scene Generation}
Our model architecture is shown in Fig.~\ref{fig:sceneformer}.  We use the transformer decoder \cite{vaswani2017attention} to generate the sequences of object properties. For each of the four properties, we train a separate model to predict the corresponding token of the current object.
Object properties are predicted in an \textit{autoregressive} manner - each object's properties are predicted conditioned on the previously predicted objects.

The distribution over the sequence $\{c_i\}$ is factorized as 
$$
p(\{c_i\};M_c) = \prod\limits_{n=1}^{N} p(c_n | o_{<n};M_c)
$$
where the category model $M_c$ expresses the distribution over the category  $c_n$ of a single object.
The factorization of the orientation sequence is
$$
p(\{\theta_i\}; M_{\theta}) = \prod\limits_{n=1}^{N}p(\theta_n | c_{\leq n}, o_{<n}; M_{\theta})
$$
The location and dimension models $M_{loc}$ and $M_{dim}$ are conditioned on tokens generated so far, for example the factorization for $\{y_i\}$ is 
$$
p(\{y_i\};M_{loc}) = 
\prod\limits_{n=1}^{N} 
p(y_n | c_{\leq n}, 
        \theta_{\leq n},
        x_{\leq n},
        o_{<n};
        M_{loc})
$$

Each model is conditioned on the output of the previous models. The category model is conditioned on all previous objects. The orientation model is conditioned on the category of the current object, as well as all other properties of all previous objects and so on. 

We find empirically that a single transformer with comparable model capacity is difficult to optimize, and tends to produce unrealistic scenes, since learning a sequence composed of different features is hard. 
Further, conditioning on the object dimensions does not improve performance, hence none of our models, except the dimension model, take the dimension sequence as input. Intuitively, a prior over object dimensions is learned and the model infers likely object dimensions of previous objects based on their categories and locations.

Similarly, swapping the order of location and orientation models led to unrealistic object locations. Learning locations is the most difficult task of the 4 properties considered and the location model benefits from more inputs.

\paragraph{Sequence Representation}
Each model takes multiple sequences as input.
Since the location $(x, y, z)$ and dimension $(l, w, h)$ of each object are 3-dimensional, the input sequences for location  and dimension are obtained by concatenating tuples of $(x_{i}, y_{i}, z_{i})_{i}$ and $(l_{i}, w_{i}, h_{i})_{i}$. Therefore, the other input sequences should be repeated 3 times. 
In order to condition on properties of the current object during training, the corresponding input sequence is shifted to the left by one token. For example, as shown in Fig.~\ref{fig:sceneformer}, the category input sequence for the orientation model is shifted towards the left by one token, so that the orientation is generated conditioned on the category of the current object.   

\paragraph{Embedding}
We use learned position and value embeddings \cite{child2019generating} of the input sequences for each model. The position embedding indicates the position in the sequence the object belongs to. The value embedding indicates the token's value. For the location and dimension models, we add another embedding layer to indicate whether the token is an $x, y$ or $z$ coordinate for location sequence, and whether the token is $l, w$ or $h$ for the dimension sequence. Then we combine the embeddings of all sequences by addition.

The output embedding is converted to $N$ logits with a linear layer, where $N$ is the number of possible quantized values. Each network is trained independently with a cross entropy loss.

\paragraph{Inference}
During inference, properties of objects are generated in the order of class category, orientation, location and dimension. Once a new token is generated, the corresponding sequence is appended with the new token and given as input to the next model. 
The location and dimension models are run three times each, to obtain three different output tokens $(x, y, z)$ and $(l, w, h)$ respectively.

We use probabilistic nucleus sampling (top-\textit{p} sampling)  on the category model outputs with $p=0.9$, and pick the token with the maximum score from the other 3 models. If any model outputs a stop token, the sequence is terminated.

\subsection{Room-Layout Conditioned Scene Generation}
We generate an indoor scene conditioned on a room layout, defining the floor, windows and doors (walls are assumed to lie at the edges of the floor). 
The floor is represented as a binary image and encoded by a series of residual blocks \cite{he2016deep}, shown in Fig.~\ref{fig:sceneformer}. We use binary images of size $512\times 512$ and obtain feature maps of size $16\times 16\times E$, where $E$ is embedding dimension of the transformer model. A discrete 2D coordinate embedding is added to this feature map, which is then flattened to obtain a $256\times E$ sequence. The SceneFormer decoder then performs cross-attention on the embedded sequence. 

The locations of doors and windows are inserted as objects at the beginning of the input sequence. Hence, inference starts with a sequence consisting of the start token along with tokens for doors and windows. Such leading tokens have also been used in works such as CTRL~\cite{keskar2019ctrl}. We experimented with doors and windows as multiple channels in the floor plan image, but locations were not recognized precisely enough, causing collisions.

We do not use an additional loss to enforce that the generated objects are within the input floor region.

\begin{figure*}[h!]
\vspace{-0.1cm}
\centering
\includegraphics[width=0.8\textwidth]{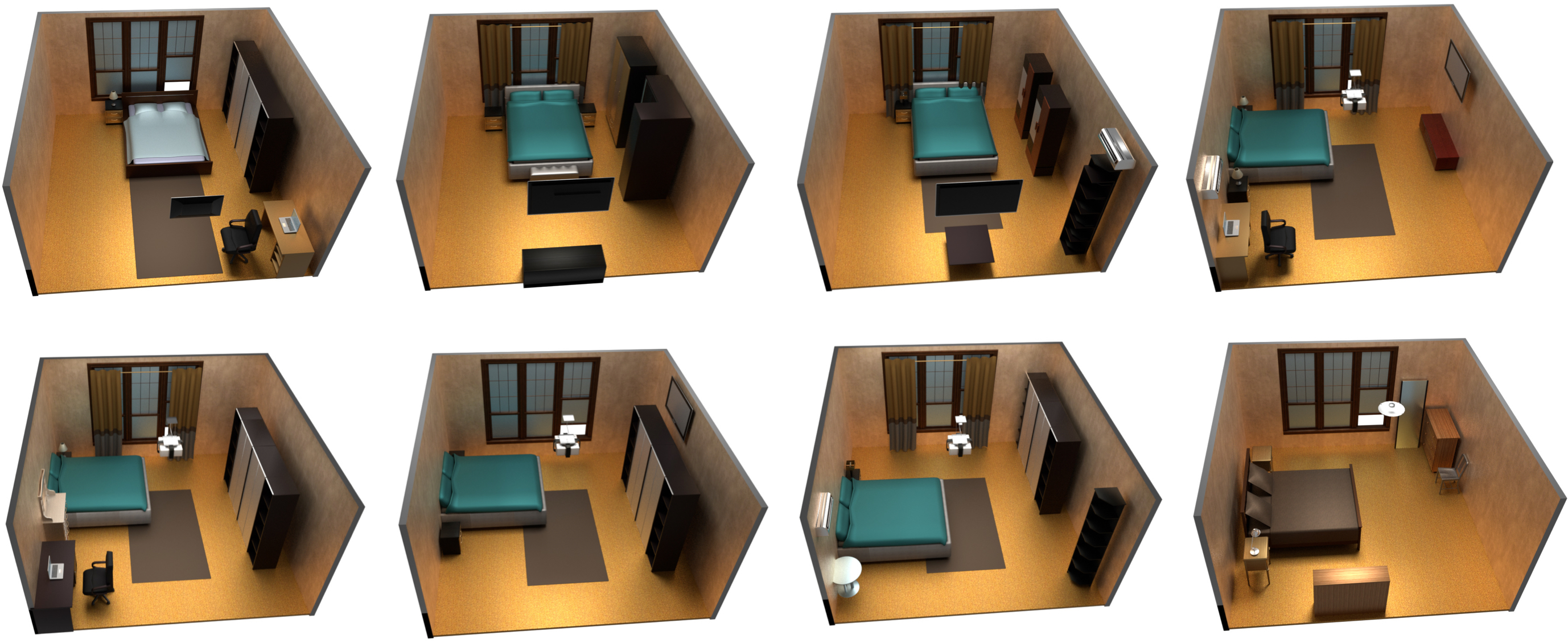}
\vspace{-0.3cm}
\caption{Scene diversity: a ground truth scene for a room layout is shown at the bottom right. The other 7 scenes are generated conditioned on its room layout. Note that tables and chairs are learned to generate (or not generate) together in combination. Location diversity is also obtained from  rotation and jitter augmentations.}
\label{fig:scene_diversity}
\end{figure*}

\subsection{Text-Conditioned Scene Generation}
In scene generation from text, a room is described by a list of sentences. 
We use the first 3 sentences, tokenize them and pad the token sequence to a maximum length of 40 tokens. We then embed each word with an embedding function. We experiment with 
GloVe \cite{pennington2014glove}, ELMo \cite{peters2018deepelmo}, and BERT \cite{devlin2018bert}. We obtain fixed-size word embeddings with dimension $d$ ($100, 1024$ and $768$ respectively) using the Flair library \cite{akbik2018coling}, and then a 2-layer MLP to convert from $d$ to $E$ dimensions, where $E$ is the dimension of the SceneFormer embedding. 
For the text-conditional model, we use decoders only for the category and location models, since our sentences only describe object classes and their spatial relations. The \textit{decoders} for orientation and dimension models are replaced by \textit{encoders} without cross-attention. 
We do not use an additional loss to align the transformer output with the input text; this relation is learned implicitly.

\subsection{Object Insertion}
For each generated object, we find the CAD model that is closest in size using the L2-norm over the dimensions of the object. If this causes a collision with 3D IoU of up to $0.05$, we reselect the next CAD model, and repeat this upto 20 times. If none of the models fit in the predicted location, we resample the object category. This heuristic is important for large objects placed in rooms with little space left.

\begin{figure*}[h!]
\centering
\includegraphics[width=1\textwidth]{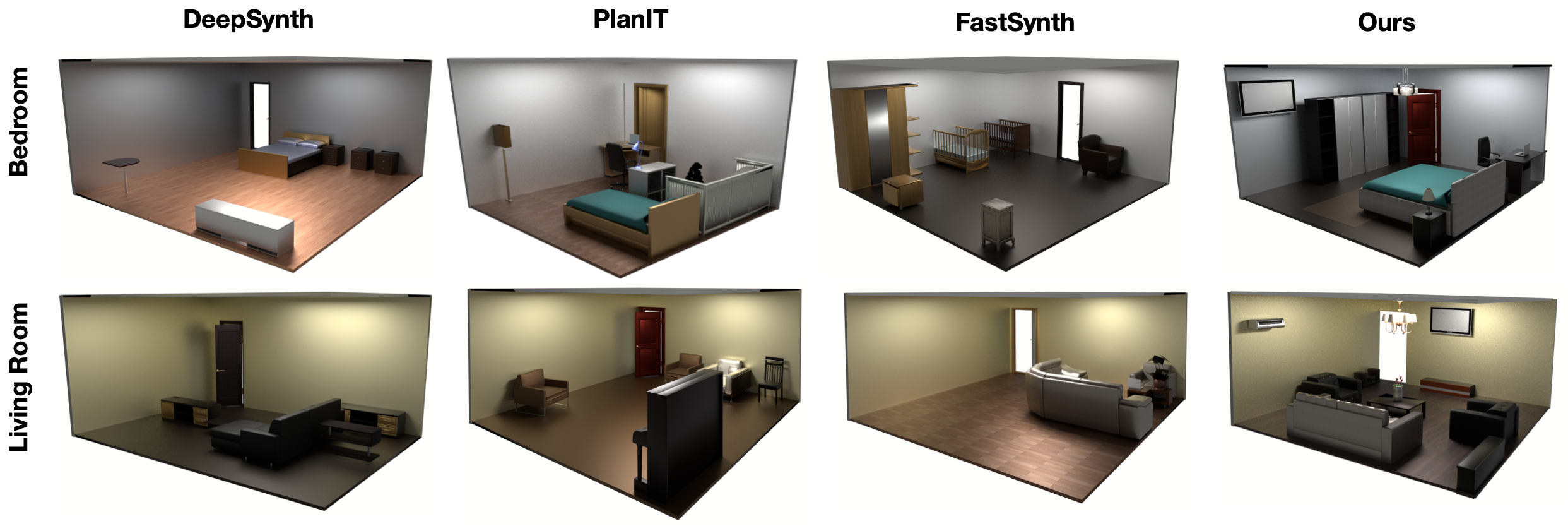}
\vspace{-0.7cm}
\caption{Bedroom and living room scenes generated by our method in comparison with state of the art. Methods are ordered from left to right in increasing realism from the perceptual study.}
\label{fig:scene_comparison}
\end{figure*}

\subsection{Data and Training Details}
We use bedrooms and living rooms from a human-created scene dataset, referred to as \textit{GT} (ground truth) \cite{song2017semantic} and filter the bad samples as done in earlier works \cite{wang2019planit, ritchie2019fast} to obtain a total of 6351 bedrooms and 1099 living rooms, which are split 80:20 into training and validation sets. 
We use 50 object categories for bedrooms and 39 object categories for living rooms.
Rooms are augmented with rotations from the set $(0, 90, 180, 270)$ degrees, and object's location jitter is sampled uniformly from $(0, 0.5)$.
We train with a learning rate of $3e^{-4}$ and apply cosine annealing with restarts after 40k iterations, for a maximum of 2000 epochs, using the Adam optimizer with a $0.001$ weight decay and a batch size of 128. All experiments are run on a single Nvidia RTX 2080 Ti. Training takes $\approx 4$ hours for each model. 

To generate textual scene descriptions, we use a heuristic method. We first extract \textit{relations} between the objects in the scene, following \cite{carion2020end}. All related objects within a distance of $2.5m$ are retained. In addition, objects can only be related to an object that appears earlier in the sequence. Then, we use a set of rules to generate sentences from these filtered relations. The first sentence mentions up to the first 3 objects in the room's category sequence $\{c_i\}$. We then iterate over all objects except the first in the sequence, and with probability $0.3$ describe an object $o_i$ using its relation to another object $o_j, j< i$ which has already been described. Each relation is a tuple $(o_i, rel, o_j)$, where $rel$ is the relation type.

\begin{figure*}[h!]
\centering
\includegraphics[width=0.9\textwidth]{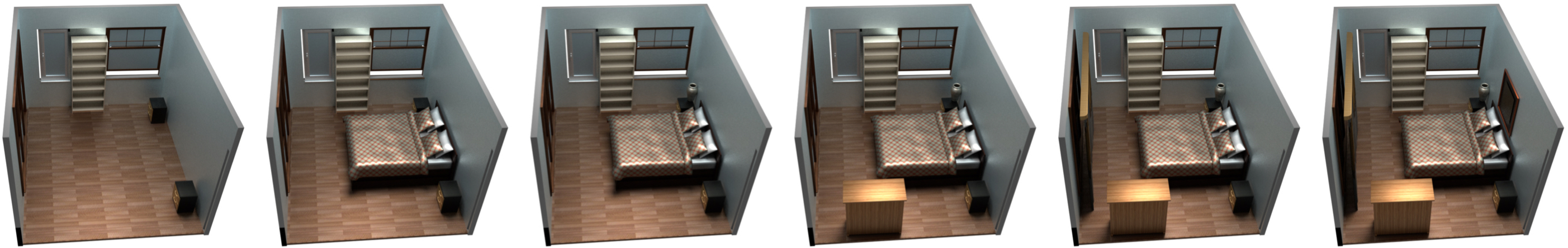}
\vspace{-0.2cm}
\caption{Scene completion sequence, showing the iterative process of adding objects; the leftmost is the input to the model, and the others are obtained sequentially by adding objects.}
\label{fig:scene_completion}
\end{figure*}

\section{Results}
\subsection{Qualitative Layout-conditioned Generation}

\paragraph{Scene Synthesis Comparison}
Figure \ref{fig:scene_comparison} shows scenes generated by our model compared to the scenes from DeepSynth~\cite{wang2018deep}, PlanIT~\cite{wang2019planit} and FastSynth~\cite{ritchie2019fast}. Our approach  generates more complex scenes in terms of object categories and object relations. Previous works perform image-based scene generation and therefore can only generate objects on the floor, or objects supported by a plane such as laptop. In contrast, we are able to generate objects on the walls (e.g., air conditioner, TV), and on the ceiling (e.g., chandelier). Object relations are also learned - the television is placed opposite the bed or sofa, curtains are placed on windows, bed stands are placed on either side of the bed, etc..

\newlength{\tempdima}
\newcommand{\rowname}[1]% #1 = text
{\rotatebox{90}{\makebox[\tempdima][c]{\textbf{#1}}}}

\begin{figure}[h!]
\vspace{-0.1cm}
\settoheight{\tempdima}{\includegraphics[width=.32\linewidth]{example-image-a}}%

\centering
\begin{tabular}{@{}c@{ }c@{ }c@{ }c@{}}

&\textbf{double bed/stand} & \textbf{double bed/tv} & \textbf{desk/chair} \\
\rowname{GT}&
\includegraphics[width=.3\linewidth]{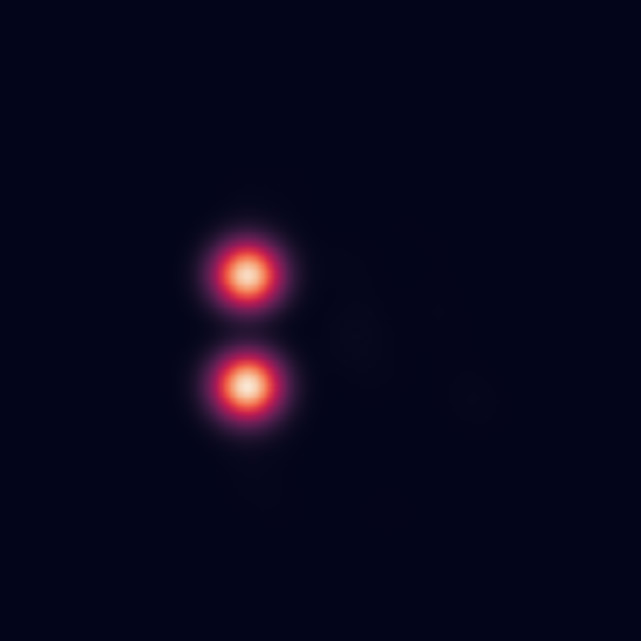}&
\includegraphics[width=.3\linewidth]{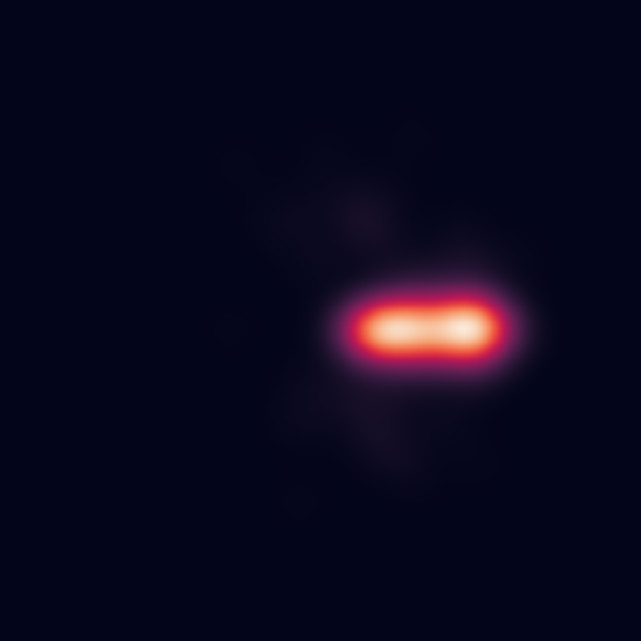}&
\includegraphics[width=.3\linewidth]{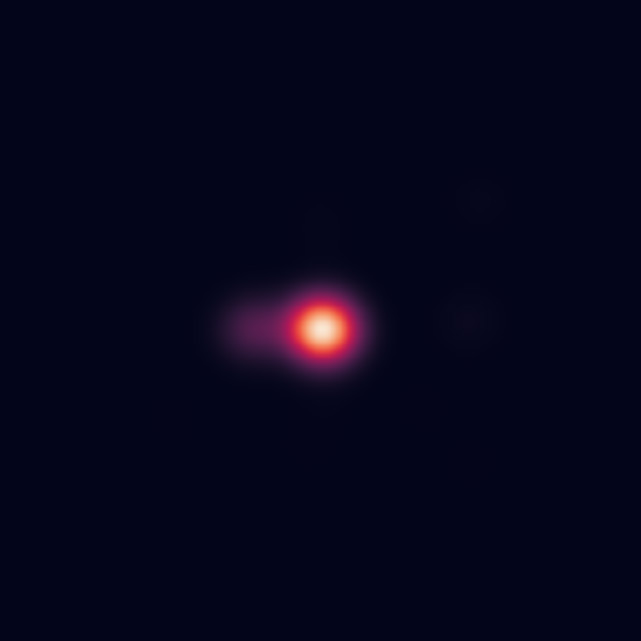}\\

\rowname{Ours}&
\includegraphics[width=.3\linewidth]{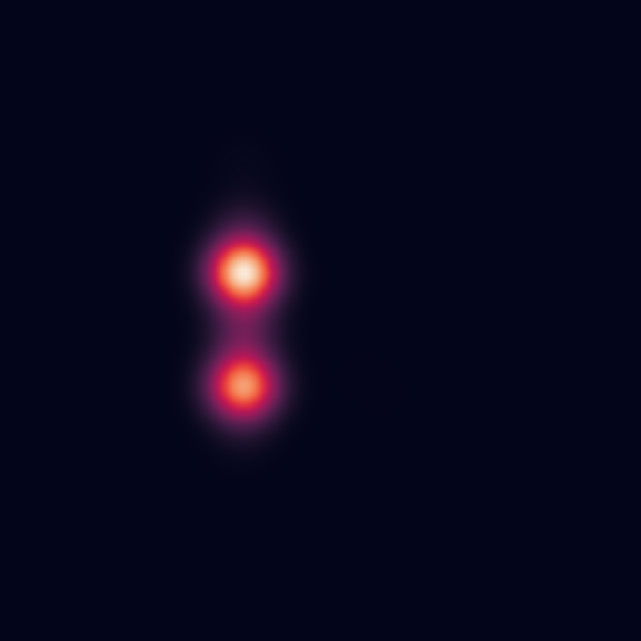}&
\includegraphics[width=.3\linewidth]{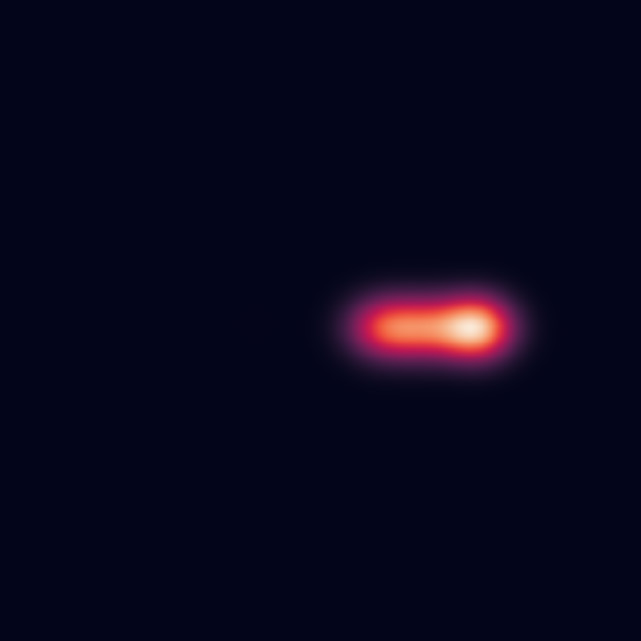}&
\includegraphics[width=.3\linewidth]{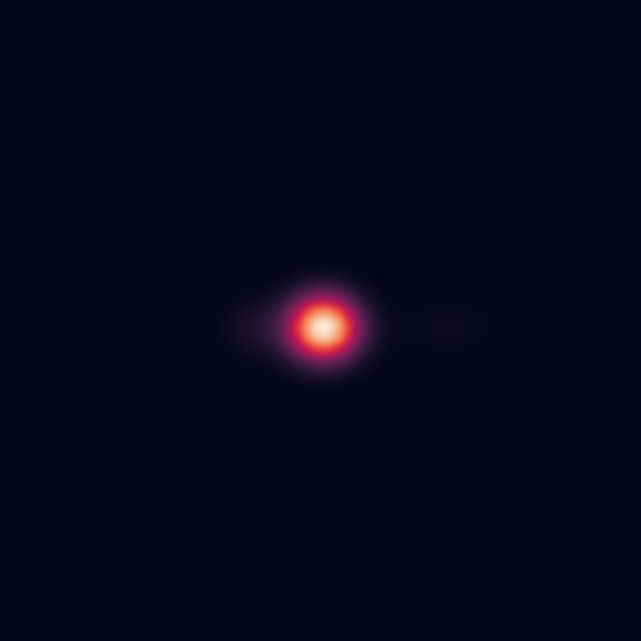}\\[1ex]

\end{tabular}
\vspace{-0.5cm}
\caption{Object location heatmap for pairs of objects in bedroom scenes, ground truth and ours. 
We learn to capture common patterns in object relations.
}
\label{fig:heap_map_bedroom}
\end{figure}

\begin{figure}[h!]

\settoheight{\tempdima}{\includegraphics[width=.32\linewidth]{example-image-a}}%
\centering\begin{tabular}{@{}c@{ }c@{ }c@{ }c@{}}
&\textbf{plant/sofa} & \textbf{sofa/coffee table} & \textbf{sofa/tv} \\
\rowname{GT}&
\includegraphics[width=.3\linewidth]{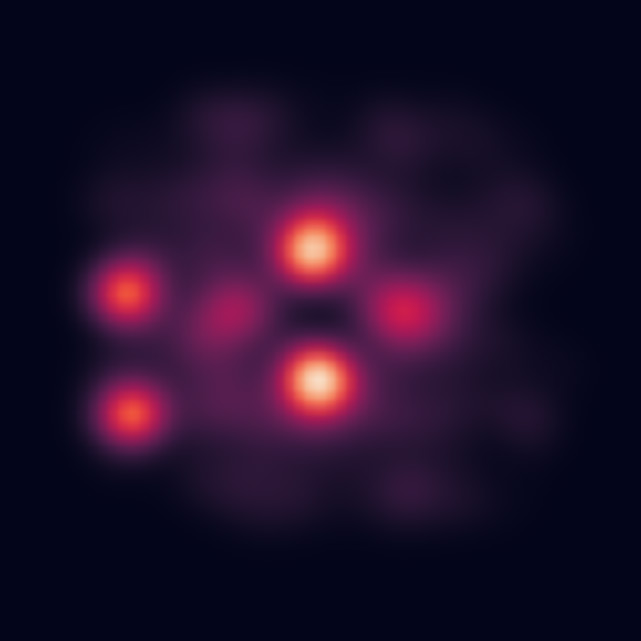}&
\includegraphics[width=.3\linewidth]{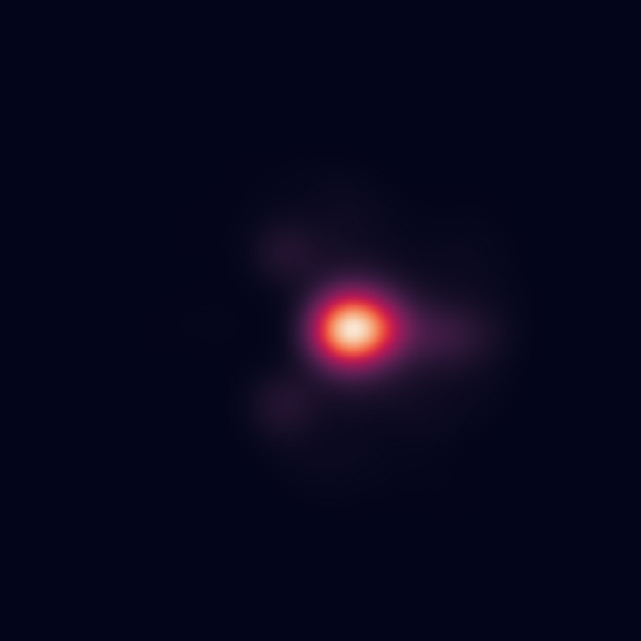}&
\includegraphics[width=.3\linewidth]{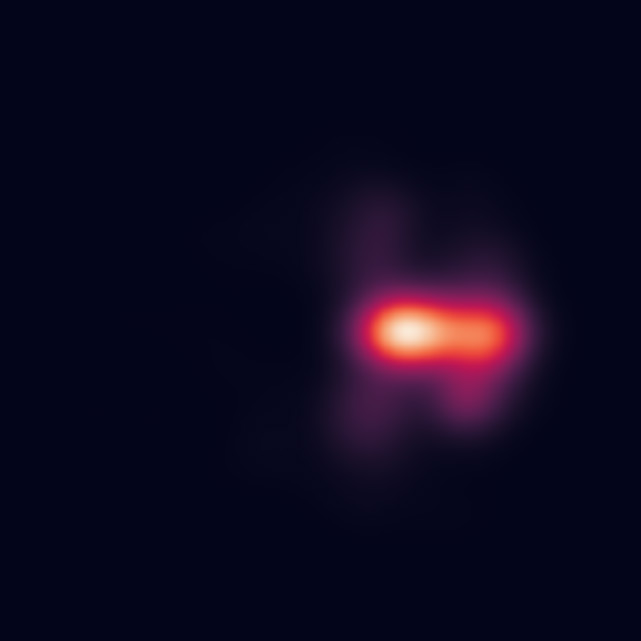}\\

\rowname{Ours}&
\includegraphics[width=.3\linewidth]{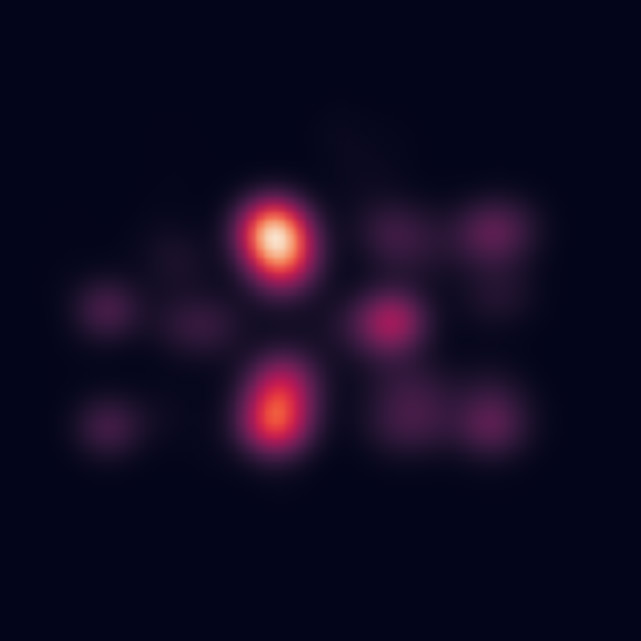}&
\includegraphics[width=.3\linewidth]{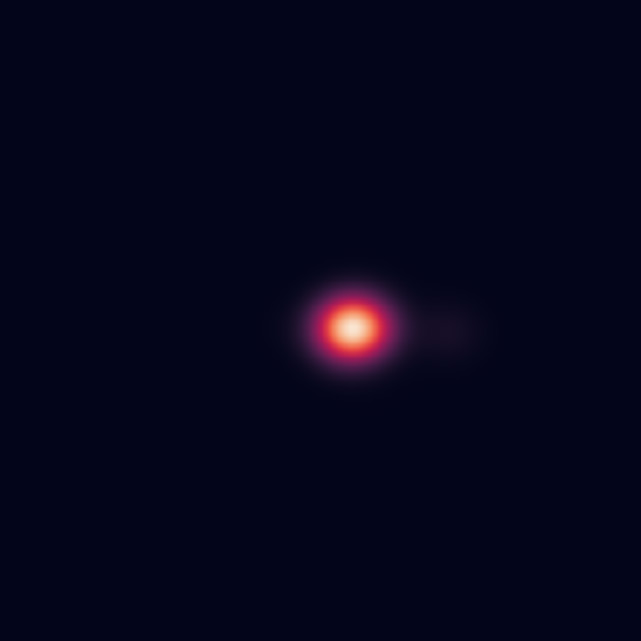}&
\includegraphics[width=.3\linewidth]{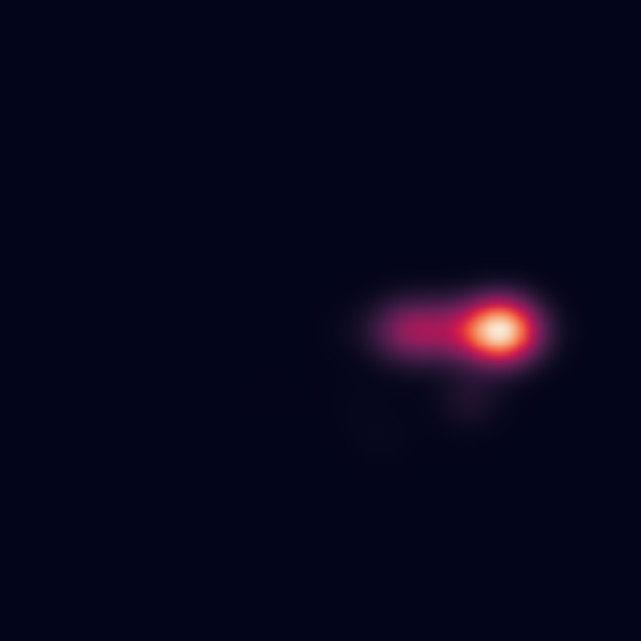}\\[1ex]

\end{tabular}
\vspace{-0.5cm}
\caption{Object location heatmap for pairs of objects in living room scenes, ground truth and ours.}
\label{fig:heap_map_living}
\end{figure}

\begin{figure}[h!]
\centering
\includegraphics[width=0.5\textwidth]{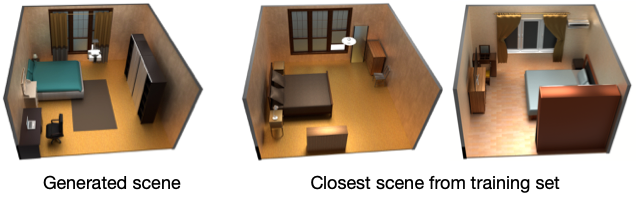}
\vspace{-0.7cm}
\caption{Closest scenes from the training set to a generated scene from our approach, based on room shape and object class, showing the ability to generate novel scenes.}
\label{fig:memorize}
\end{figure}

\smallskip 
\noindent \textbf{Scene Completion}
We show the effectiveness of our method on the scene completion task in Fig.~\ref{fig:scene_completion}. Given an incomplete scene with a few objects, our model can add relevant and missing objects to complete the scene. As a result of training on sorted sequences, the model first generates large and frequent objects before generating small and infrequent objects in a greedy manner. Thus, stopping generation at an intermediate stage results in a realistic scene, and a user can potentially interactively choose how complete the scene must be.

\smallskip 
\noindent \textbf{Scene Diversity and Memorization} 
Various generated scenes conditioned on the same input room shape are shown in Fig.~\ref{fig:scene_diversity}. We are able to generate different sets of objects and object orientations that are consistent with each other. To show our model does not simply memorize the training samples, we compare the generated scene with the nearest neighbor in the training set based on two different methods: room shape and object class, as shown in Fig.~\ref{fig:memorize}. This indicates our ability to generate novel scenes.

\smallskip 
\noindent \textbf{Object Category Heatmap}
Figs.~\ref{fig:heap_map_bedroom} and \ref{fig:heap_map_living} show heatmaps of relative locations between objects, demonstrating that our model effectively captures object relations from the data. 

\subsection{Qualitative Text-conditioned Scene Generation}
Text-conditioned scenes generated by our model and the corresponding descriptions are shown in Fig.~\ref{fig:text_scene}. Generated scenes largely respect the input sentence in terms of object categories and relations, and have additional objects to improve the completeness of the scene.

\begin{figure*}[!h]
\vspace{-0.4cm}
\centering\begin{tabular}{@{}c@{ }c@{ }c@{ }c@{ }}
\captionsetup{justification=centering}
\subfloat[There is a wardrobe cabinet, a double bed and a desk. There is an office chair next to the desk.
]{\includegraphics[width=.3\linewidth]{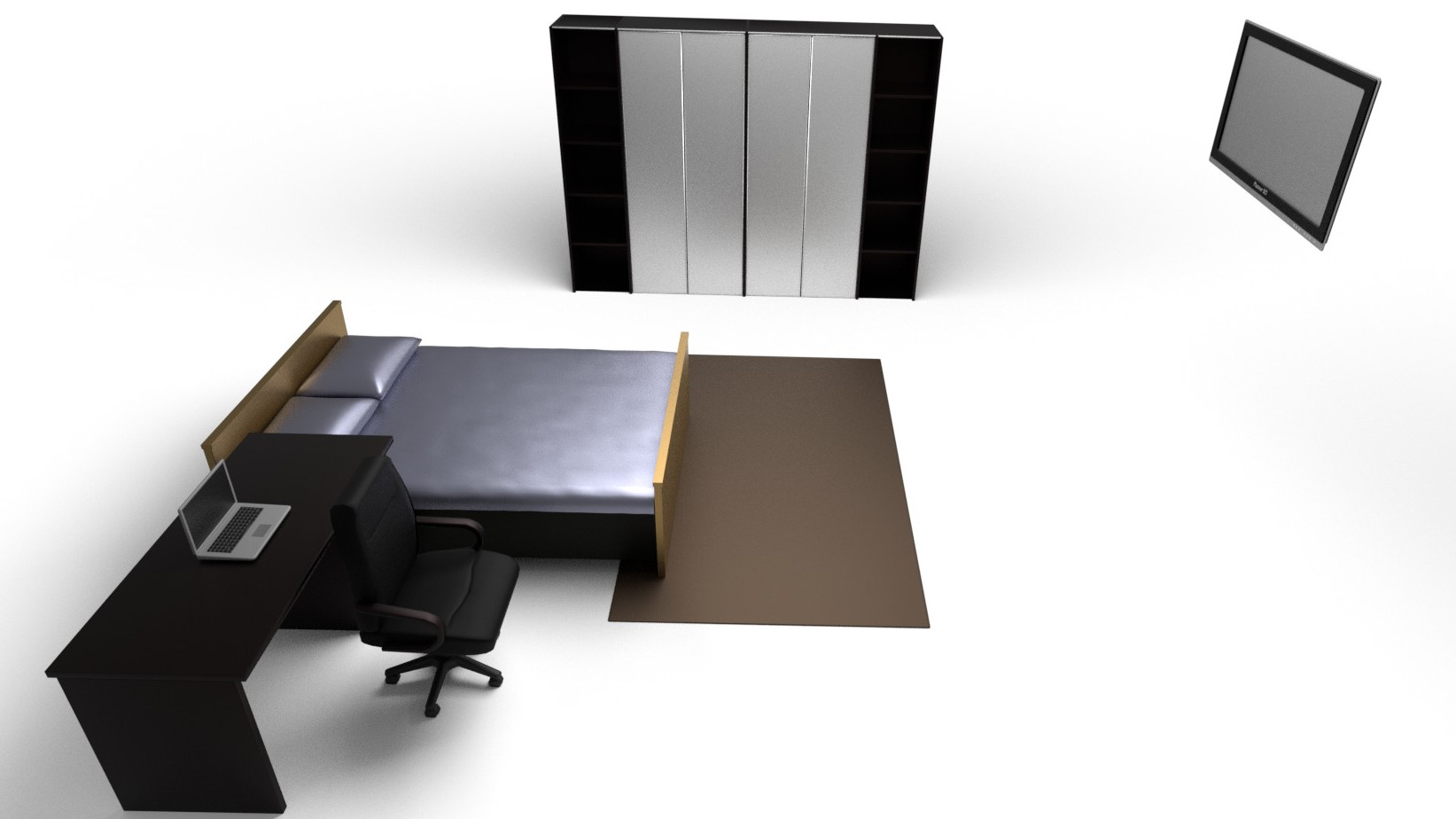}} &

\captionsetup{justification=centering}
\subfloat[The room contains a bunker bed and a wardrobe cabinet.
]{\includegraphics[width=.3\linewidth]{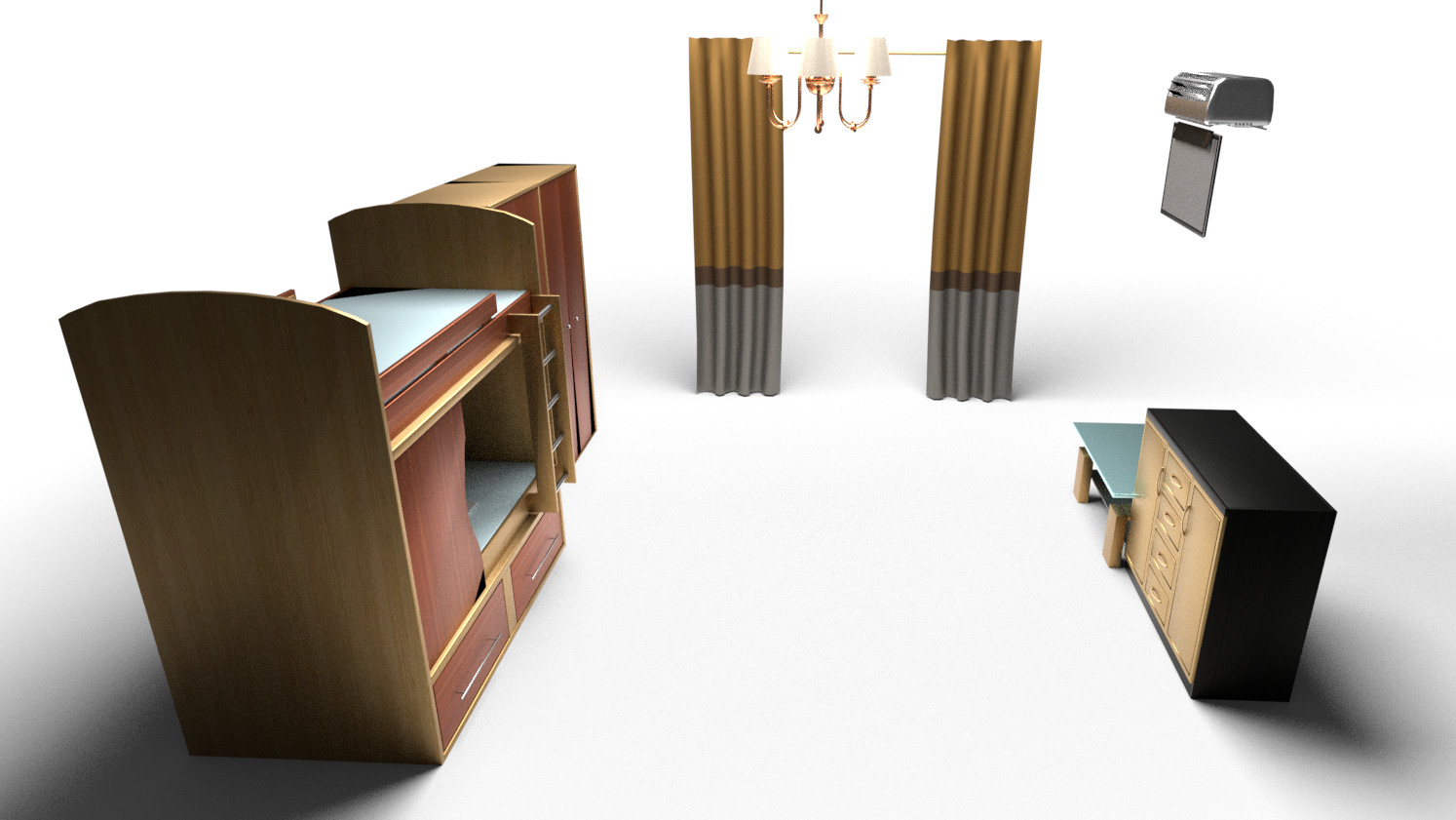}} &

\captionsetup{justification=centering}
\subfloat[The room has a wardrobe cabinet, a stand and a double bed.
]{\includegraphics[width=.3\linewidth]{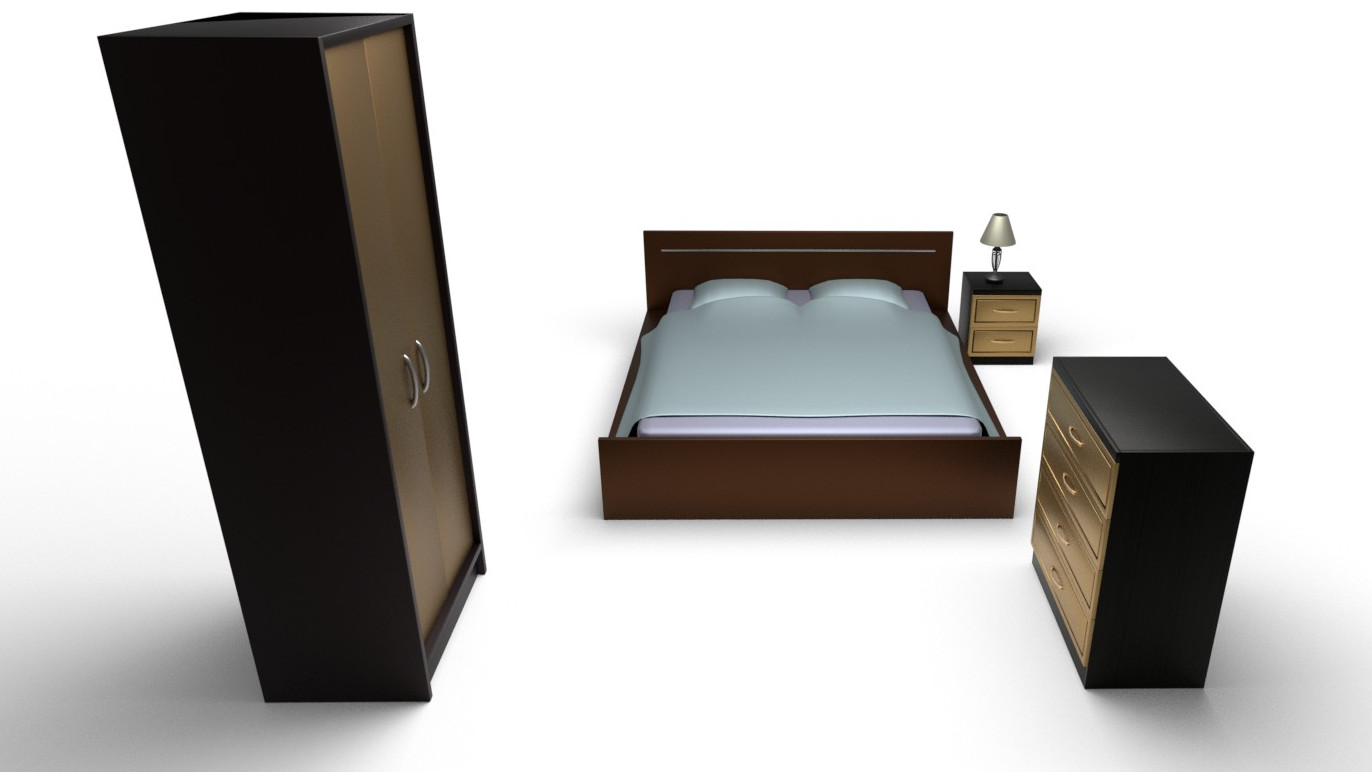}} \\

\captionsetup{justification=centering}
\subfloat[In the room we see a heater and a wardrobe cabinet. There is a stand to the right of the wardrobe cabinet.
]{\includegraphics[width=.3\linewidth]{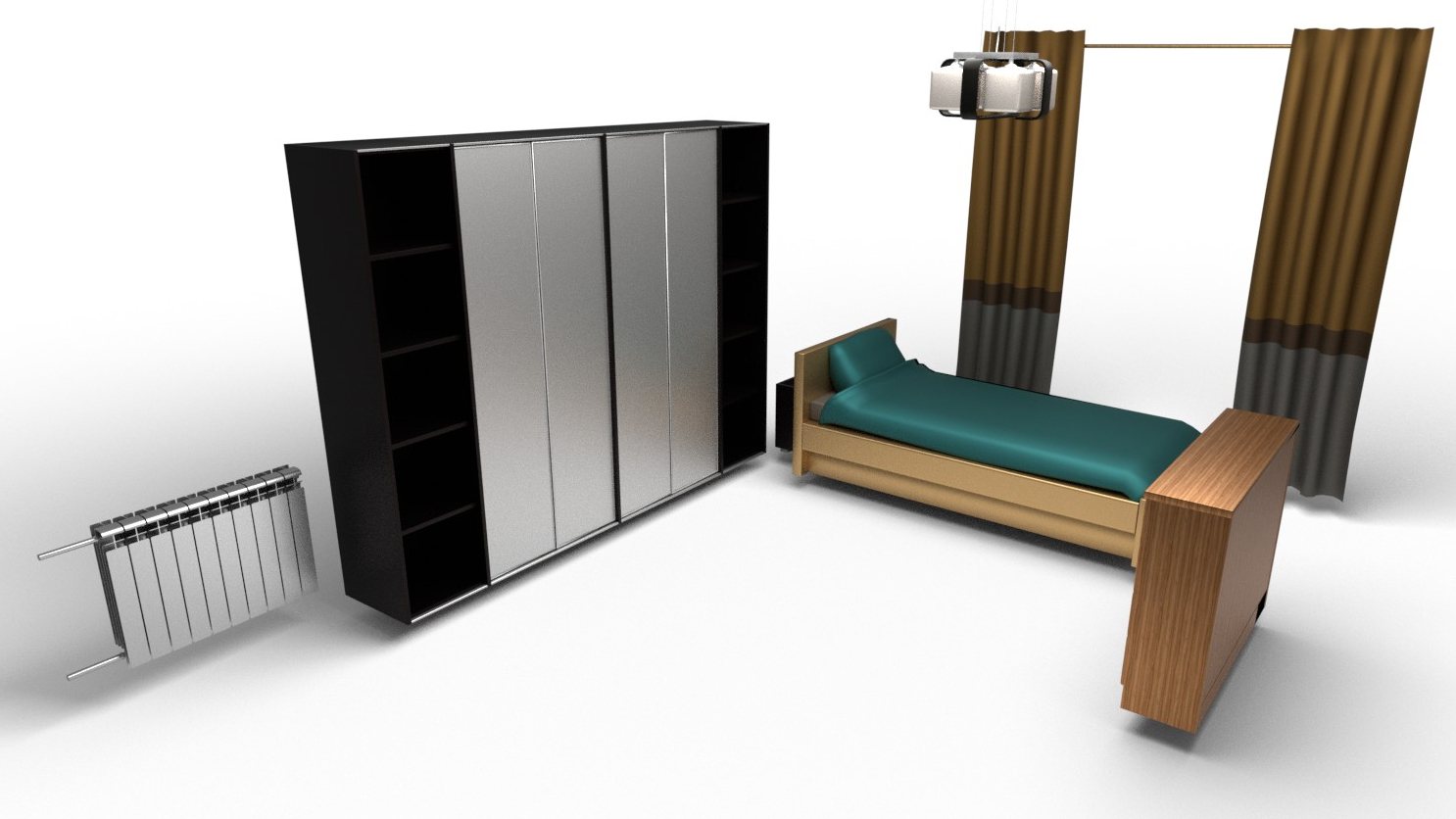}} &

\captionsetup{justification=centering}
\subfloat[In the room we see two ottomans. The wardrobe cabinet is to the left of the second ottoman.
]{\includegraphics[width=.3\linewidth]{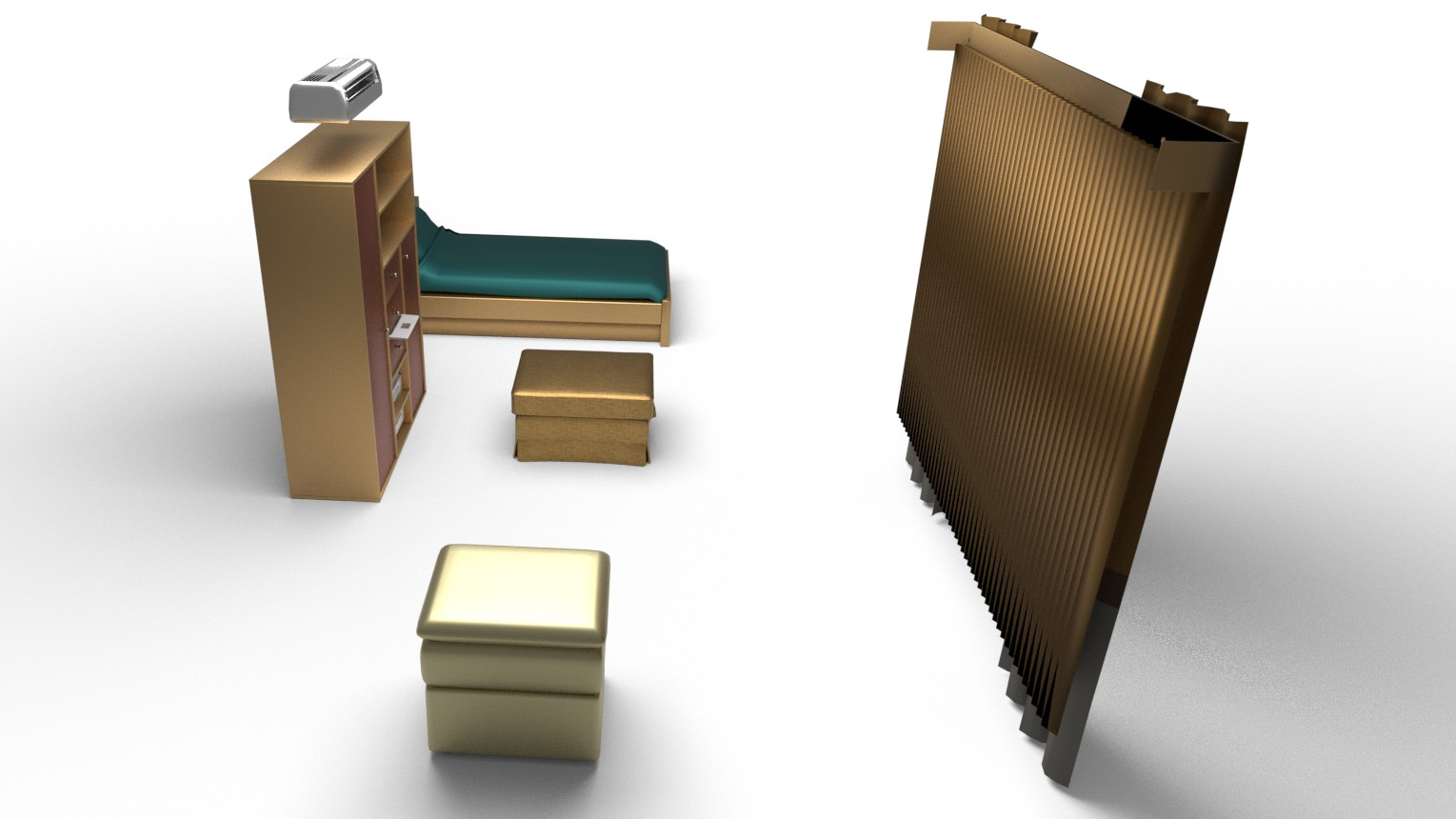}} &

\captionsetup{justification=centering}
\subfloat[This room has a wall lamp, a heater and a wardrobe cabinet. There is a double bed to the left of the heater.
]{\includegraphics[width=.3\linewidth]{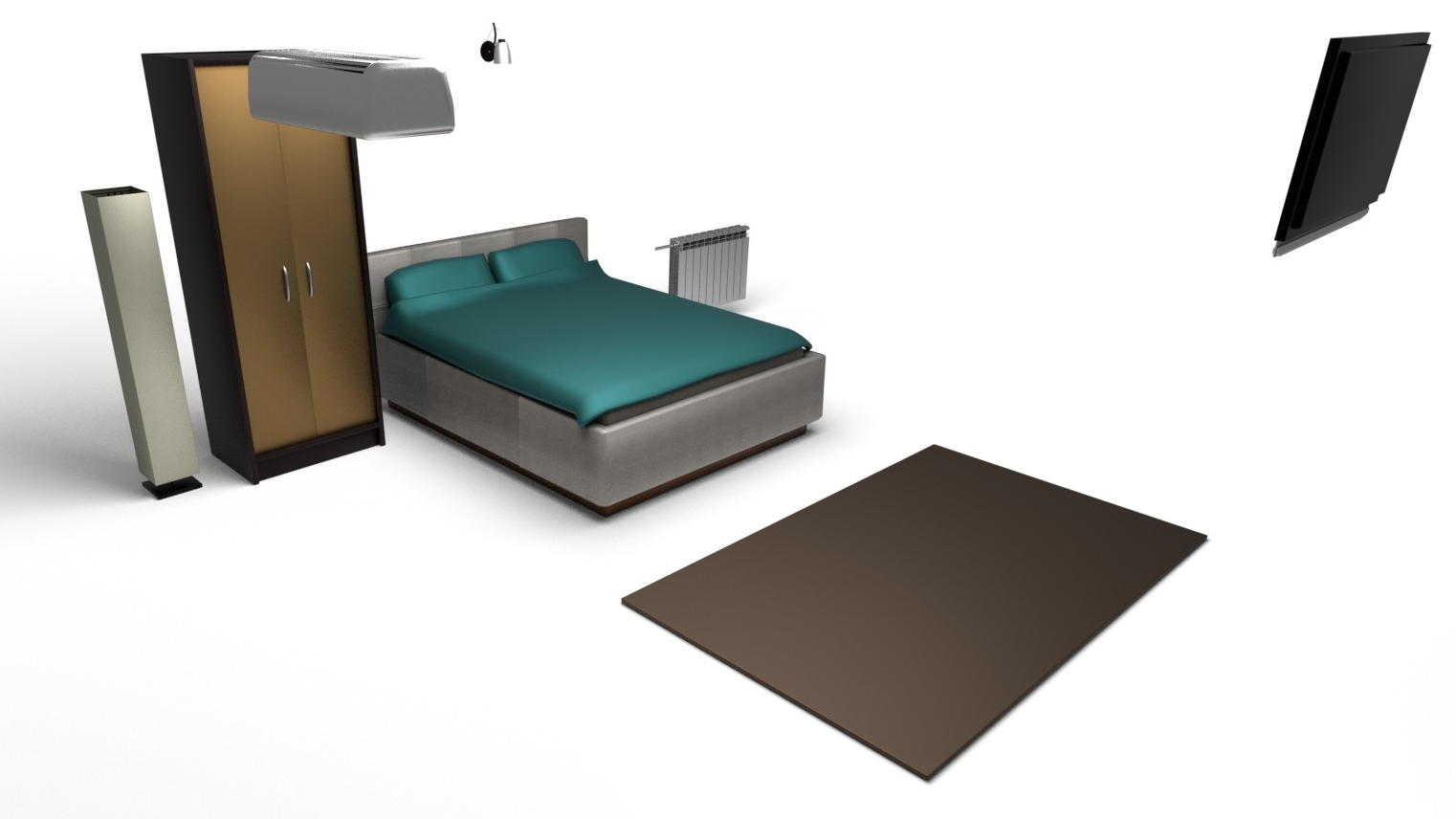}}

\end{tabular}
\vspace{-0.2cm}
\caption{Text-conditioned generated bedroom scenes shown with their corresponding text inputs. Since the room layout is not given as input, objects are placed within a layout prior learned from the ground truth scenes. Important relations such as table-chair, bed-television and sofa-television are captured by the model.}

\label{fig:text_scene}
\end{figure*}

\subsection{Quantitative Layout-conditioned Generation}

\smallskip 
\noindent \textbf{Perceptual Study}
We conducted perceptual studies, following those done by earlier works \cite{wang2019planit, ritchie2019fast}.
In each study, we compare a scene generated by our method with a scene from one of DeepSynth~\cite{ritchie2019fast}, FastSynth~\cite{ritchie2019fast}, PlanIT~\cite{wang2019planit} or from the \textit{GT} dataset \cite{song2017semantic}. In each study, users are shown 50 pairs of images - one from ours and one from the other method. The user must select the more realistic scene. We use 5 pairs of images as vigilance tests. The responses of users who do not pass all the tests are discarded.  Each study was taken by 30 users through Amazon Mechanical Turk. Tab.~\ref{table:userstudy} shows the results of this study. Our scenes are consistently preferred over other methods, ranging from 53.5\% up to 65\%  preference over other methods. 

\begin{table}[h!]
\centering
\begin{tabular}{l @{\hspace{\tabcolsep}} c @{\hspace{\tabcolsep}} c @{\hspace{\tabcolsep}} c @{\hspace{\tabcolsep}} c @{\hspace{\tabcolsep}} } 
\hline
 Scene Type & DeepSynth &   FastSynth &  PlanIT & GT \\ [0.5ex]
 \hline
 Bedroom & $55.4  \scriptstyle \pm 6.3$ & $53.9 \scriptstyle  \pm 6.5$ & $53.5 \scriptstyle  \pm 5.1$ &  $50.2 \scriptstyle  \pm 6.6$  \\
 Living & $56.6 \scriptstyle  \pm 4.0$ & $56.7 \scriptstyle  \pm 6.1$ & $65.0 \scriptstyle  \pm 7.6$ & $48.5 \scriptstyle  \pm 6.6$  \\ 
 
\end{tabular}
\vspace{-0.3cm}
\caption{Perceptual user study of our scenes vs. GT and other methods (top row). Metric is the percentage of image pairs where our scene is preferred. Separate studies are conducted for bedroom and living room scenes. }
\label{table:userstudy}

\end{table}

\begin{figure}[h!]
\centering
\includegraphics[width=0.5\textwidth]{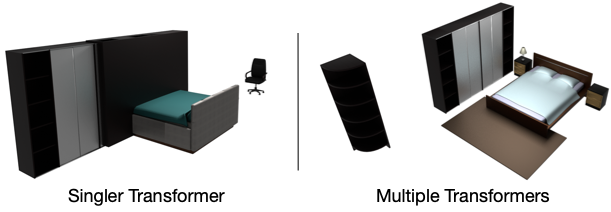}
\vspace{-0.6cm}
\caption{Scenes generated by a single transformer vs. multiple transformer models.}
\label{fig:single-multi}
\end{figure}

\paragraph{Ablation Studies}
We evaluate our design choices through ablation studies shown in Tab.~\ref{table:ablation}, using the accuracy of the next token over a fixed validation set of bedrooms. We compare the \textit{Single} model setting (1 transformer with 1 input sequence) against the \textit{Multiple} model setting (multiple transformers with parallel input sequences), and the effect of using rotation and jitter augmentation. The accuracy of the multiple model setting is the average of all four models. Augmentation gives small improvements in accuracy, while using multiple transformers leads to an improvement of 14.5 in accuracy. This is seen in the generated scenes as well in Fig.~\ref{fig:single-multi}.

\begin{table}[h!]
\centering
\begin{tabular}{l  c  c  } 
\hline
 Model &  Accuracy \\ [0.5ex]
 \hline
 Single-no augmentation & 33.1   \\
 Single-rotation & 35.4   \\
 Single-rotation \& jitter & 37.0 \\
 Multiple-rotation \& jitter & \bftab 51.5 \\
\end{tabular}
\vspace{-0.2cm}
\caption{Next token prediction accuracy for ablations. }
\label{table:ablation}
\end{table}

\smallskip
\noindent \textbf{Timing.}
We evaluate the inference time of our layout-conditioned model compared with state-of-the-art in Tab.~\ref{table:speed}. Our model is an order of magnitude faster than PlanIT and Deep Synth, and $20\%$ faster than Fast \& Flexible.

\begin{table}[!h]
\centering
\begin{tabular}{l c} 
 \hline
 Method & Time (s)  \\ [0.5ex] 
 \hline
 DeepSynth \cite{wang2018deep} & 240 \\ 

 Fast \& Flexible \cite{ritchie2019fast} & 1.85  \\
 
 PlanIT \cite{wang2019planit} & 11.7  \\ 
 
 Ours & \textbf{1.48}  \\ 
 
\end{tabular}
\vspace{-0.3cm}
\caption{Comparison of inference time.}
\label{table:speed}

\end{table}

\subsection{Quantitative Text-conditional Generation}

\smallskip
\noindent \textbf{Category Accuracy.}
We measure the fraction of objects mentioned in the scene description that are present in the generated scene, shown in Tab.~\ref{table:text-cat-acc}. We compare against 2 baselines: \textit{Uniform}, by sampling uniformly from all categories and the stop token, and \textit{GT} by sampling from the ground truth object category distribution, with the stop token having the average frequency of all objects. 

\begin{table}[h!]
\centering
\begin{tabular}{l c c } 
 \hline
 Model & Bedroom & Living Room \\ [0.5ex] 
\hline
 Uniform & 23.70 & 27.89  \\ 
 GT & 38.64 & 44.11 \\
Ours (GloVe) & 58.10 & 59.35  \\
Ours (ELMo) & 76.59 & 63.31 \\
Ours (BERT) & \textbf{84.83} & \textbf{68.81} \\

\end{tabular}
\vspace{-0.3cm}
\caption{Category accuracy of text-conditioned scene generation (percentage).}
\label{table:text-cat-acc}

\end{table}

\smallskip
\noindent \textbf{Spatial Accuracy.}
We compute the relation accuracy as the fraction of generated relations that were present in the input scene description, shown in Tab.~\ref{table:text-rel-acc}. We compare with the \textit{GT} baseline where the probability of every pair of objects being related is computed over the training set.

\begin{table}[h!]
\centering
\begin{tabular}{l c c } 
 \hline
 Model & Bedroom & Living Room \\ [0.5ex] 
\hline
GT     & 10.67 & 7.35  \\ 
Ours (GloVe) & 35.41 & 32.65  \\
Ours (ELMo)  & 39.74 & 37.50  \\
Ours (BERT)  & \textbf{46.42} & \textbf{38.46} \\

\end{tabular}
\vspace{-0.3cm}
\caption{Relation accuracy of text-conditioned scene generation (percentage).}
\label{table:text-rel-acc}
\end{table}

\smallskip 
\noindent \textbf{Perceptual Study.}
We perform a perceptual study, showing $40$ users $34$ pairs of text and images each, asking them to answer two questions for each pair: how realistic the generated scene is, and how closely the generated scene matches the input text, responding on a scale of 1 (poor) to 7 (good). Our bedroom scenes obtained a realism score of $4.61 \pm 1.84$, and a match score of $4.38 \pm 1.73$, showing that our text-conditional model generates realistic scenes while capturing the objects and relations mentioned in the text description. 

\section{Conclusion and Future Work}
We presented SceneFormer, which leverages a combination of transformer models to generates realistic indoor scenes.
SceneFormer enables flexible learning from data, implicitly learning object relations, and performing fast inference. This can enable interactive scene generation from partial scenes. 
Our model can serve as a general framework for scene generation: a different task can be solved by changing the set of object properties or conditioning inputs. Future work can consider the 3D mesh of each object to obtain global style consistency. Our model could also be used for 3D reconstruction of an indoor scene given its 3D scan as a prior. Since we perform room layout and text conditioning separately, it would be interesting to see scene generation jointly conditioned on both inputs.

\section{Acknowledgement}
This work was supported by a TUM-IAS Rudolf Mößbauer Fellowship, the ERC Starting Grant Scan2CAD (804724), and the German Research Foundation (DFG) Grant Making Machine Learning on Static and Dynamic 3D Data Practical.

{\small
\bibliographystyle{ieee_fullname}
\bibliography{main}
}

\clearpage
\appendix

\section*{Appendix}

\section{Model Details}
Further details on our model are given below. 

\paragraph{Transformers}
We use these common hyperparameters across all 4 models, unless otherwise specified. 

\begin{itemize}
    \item Maximum number of objects in the scene: 50
    \item Embedding dimension: Dimension of input and output of the transformer, 256 (1024 in location model)
    \item Dimension of transformer activations: 256 (1024 location model) 
    \item Number of transformer attention heads: 8
    \item Number of transformer blocks: 8
\end{itemize}

\paragraph{ResNet for Shape Conditioning}
We use a ResNet with 
\begin{itemize}
    \item 1 input channel
    \item 3 blocks, having 3, 3, and 4 layers respectively
    \item output dimension 256
\end{itemize}

\section{Shape-conditioned Model}

\subsection{Results}
\begin{figure*}
\centering
\includegraphics[width=0.9\textwidth]{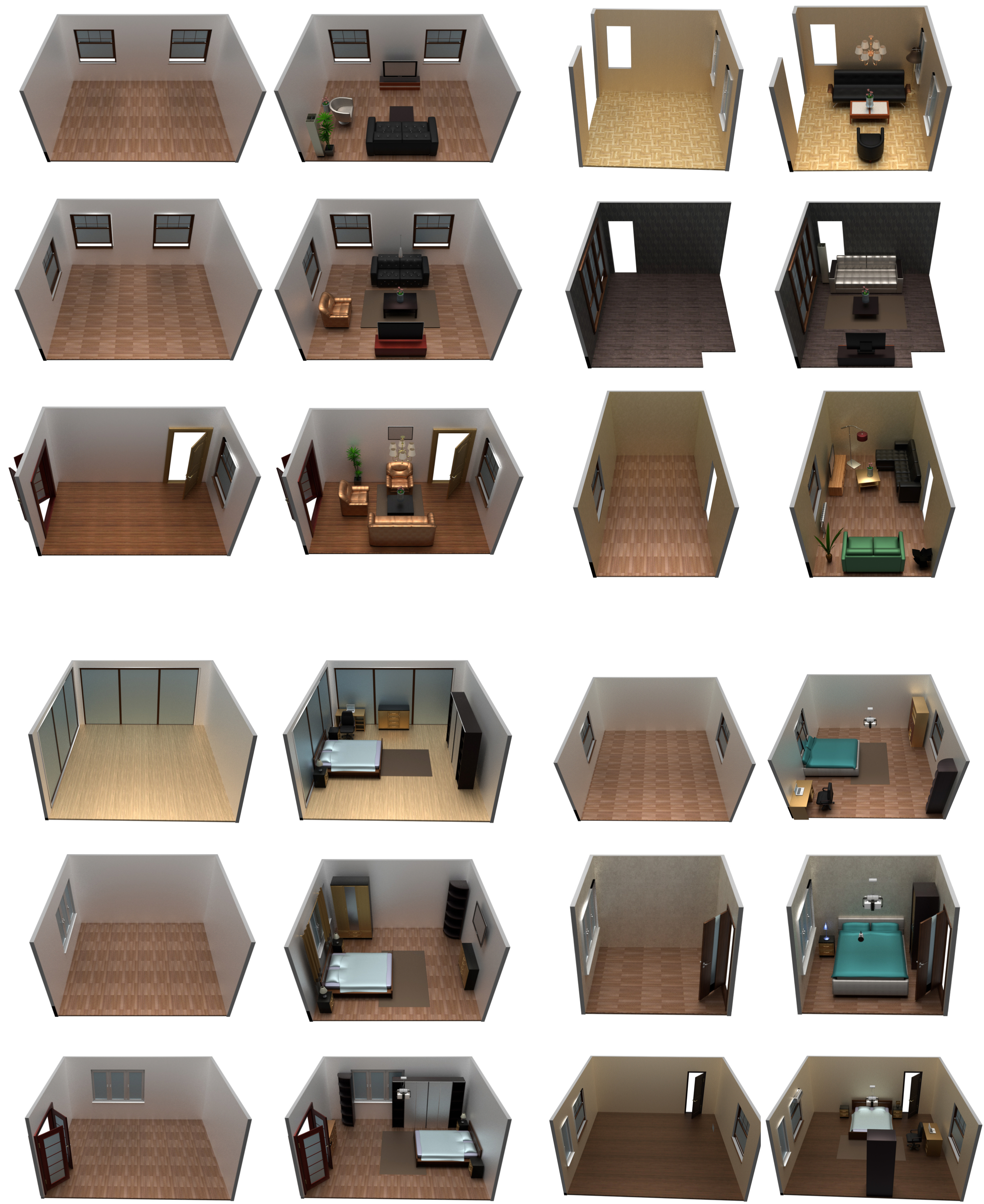}
\caption{Bedrooms and living room scenes generated by our method. In each pair, the left image is the input with the shape of the room and locations of doors and windows. The right image is the generated scene. All objects are generated within the boundaries of the room, such that they do not overlap doors and windows. In addition, meaningful relations between different object categories are learnt by the model.}
\label{fig:shapecond}
\end{figure*}

Additional bedroom and living room scenes generated by our shape-conditioned model are shown in Fig.~\ref{fig:shapecond}, along with the input room shape.
Our scenes are compared with those from DeepSynth, conditioned on the same room shapes in Fig.~\ref{fig:scene_comparison}.

\begin{figure*}
\centering
\includegraphics[width=0.9\textwidth]{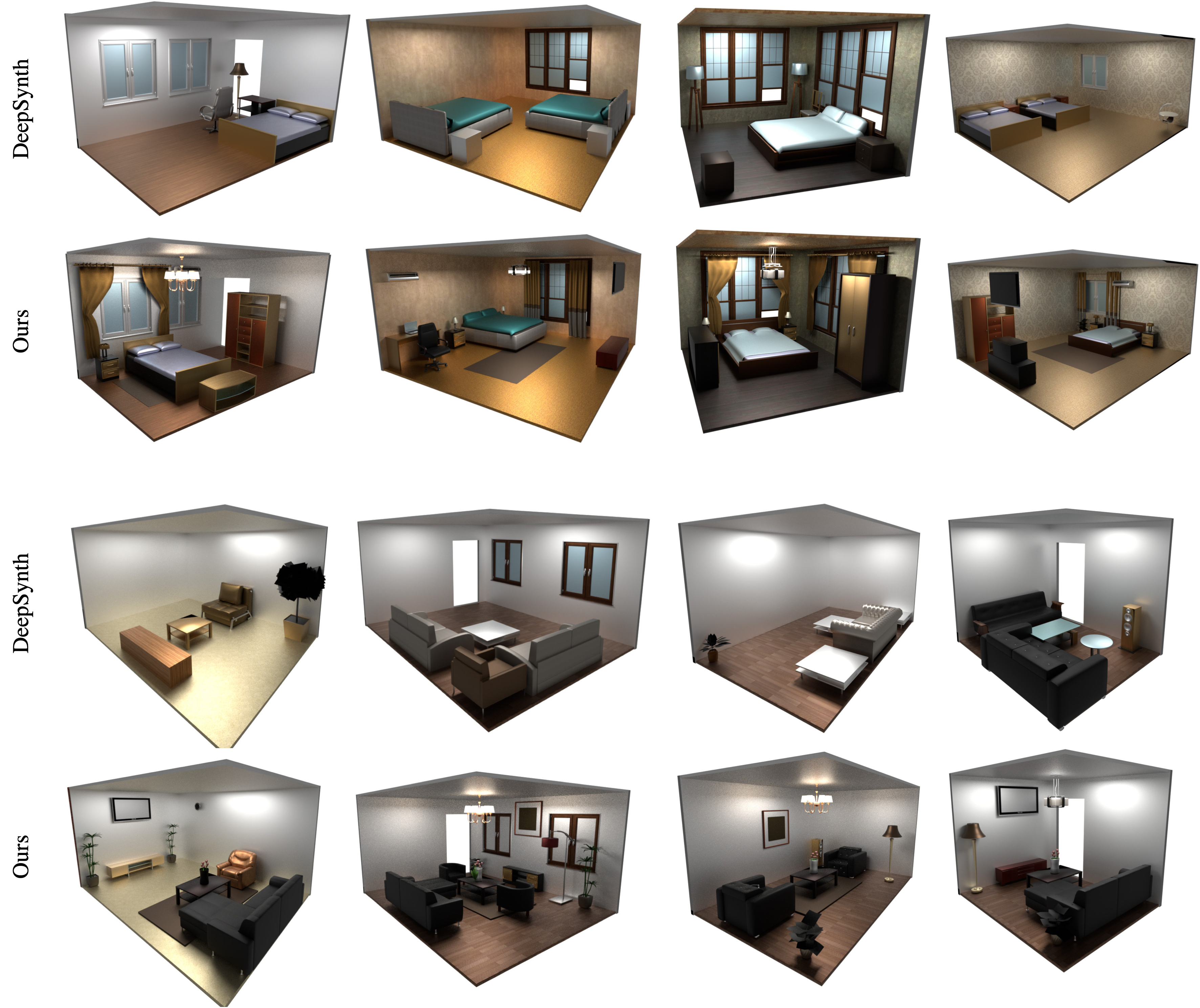}
\caption{Bedroom and living room scenes generated by DeepSynth and our method, conditioned on the same room. Our scenes have more complexity and completeness in terms of object placement.}

\label{fig:scene_comparison}
\end{figure*}

\subsection{Object Category Heatmaps}
\begin{figure*}
\centering
\includegraphics[width=0.8\textwidth]{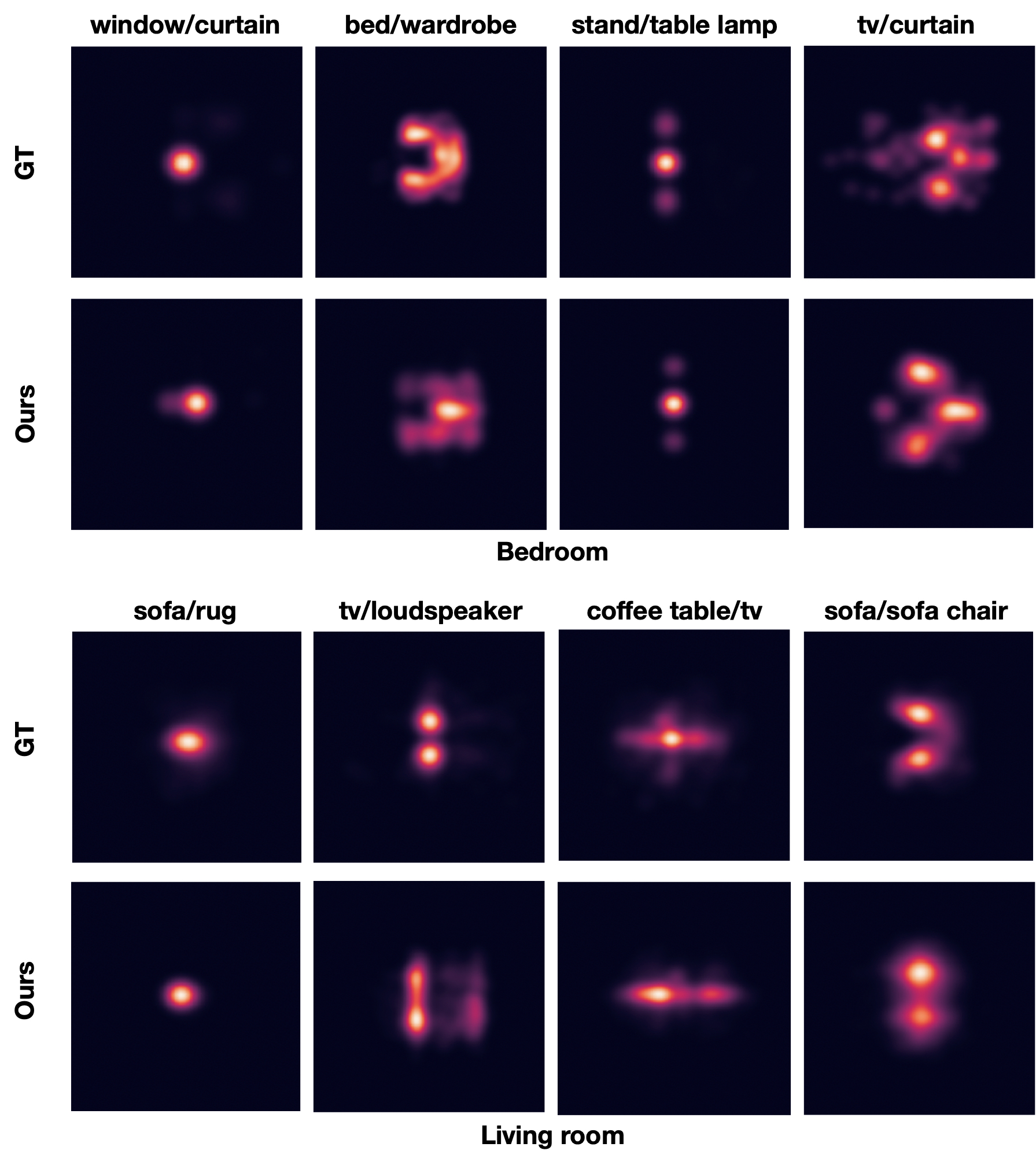}
\caption{Heatmap of object pair locations in ground truth and generated scenes. Top: pairs of object categories in bedroom scenes, bottom: living room scenes.}

\label{fig:heatmap}
\end{figure*}

Additional heatmaps comparing our generated scenes with those from the ground truth are shown in Fig.~\ref{fig:heatmap}. Our model captures the GT distribution well, while adding novel relations in some cases. This is crucical for generating diverse scenes and avoiding overfitting.

\subsection{Survey Interface}
\begin{figure*}
\centering
\includegraphics[width=0.9\textwidth]{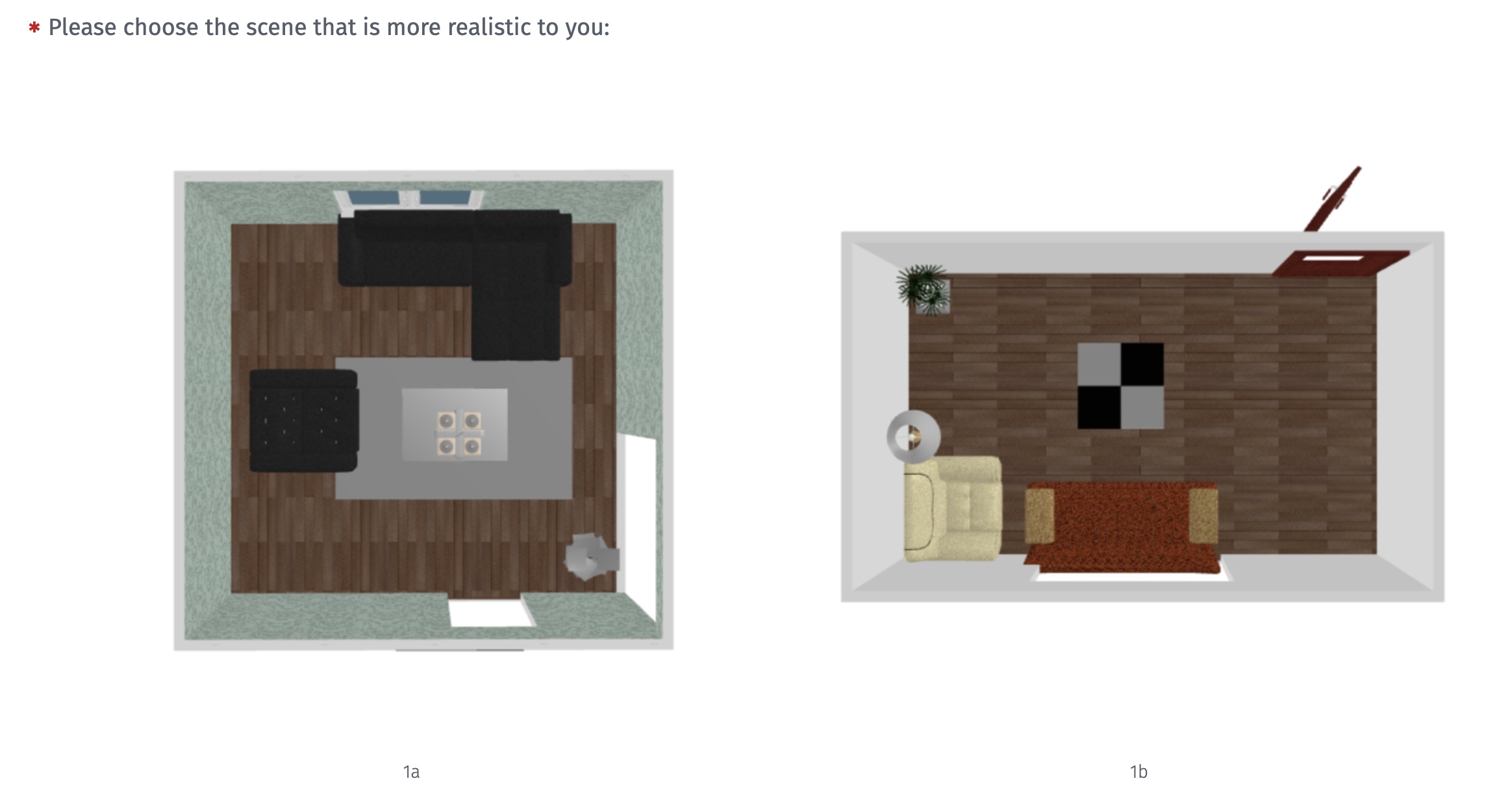}
\caption{A representative survey question from the survey for shape-conditioned images. Each question contains a pair of images. One is from our method, the other is from the dataset or generated by DeepSynth. The user has to choose the one which looks more realistic. Here we use the top down view of the scene as done in earlier works.}

\label{fig:interface}
\end{figure*}

The web interface used to conduct the survey on user preference between our scenes, and scenes from DeepSynth/PlanIT/FastSynth or from the dataset, is shown in Fig.~\ref{fig:interface}.

\section{Text-conditioned Model}
\subsection{Scene Description Generation}
We describe the method used to generate scene descriptions.

\paragraph{Relation Generation}
First, we generate a set of relations between objects in the scene. Given a sequence of objects $\{o_1, o_2, \ldots\}$, relations $r_{ij}$ are generated such that $i < j$. Based on the bounding boxes of $o_i$ and $o_j$, the relation type is classified as one of these - \textit{on, above, surrounding, inside, right of, left of, behind, in front of}. At the same time, the distance between the bounding box centers is computed.

\paragraph{Description Generation}
The first 2 or 3 objects in the sequence are described in the first sentence, this choice is made uniformly. We use a fixed set of starting phrases for the sentence -- \textit{The room has, In the room there are, The room contains, This room has, There are, In the room we see}, and then add a list of the first 2 or 3 objects. Repeated objects are mentioned as the count followed by the object category.

The remaining sentences describe objects in relation to any object that has already been described. We iterate through objects that have not been described, and choose to describe any of them with a probability of $0.7$. Then, we pick one of the relations of this object to an object that has already been described, such that the distance between them is less than the threshold of $2.5$ meters. This choice of relation is done uniformly.

Similar to the earlier procedure, ordinal prefixes such as \textit{second} and \textit{third} are added when the new object is not the first in its category to be described. In addition, objects of the same category are not related together; we reject sentences such as ``There is a second table next to the first table''. 

Next, we choose the appropriate article for each object (\textit{a, an, the)}, and choose a sentence template based on the relation between the objects, such as -- \textit{There is a table next to the chair}, \textit{There is a sofa to the right of the wardrobe cabinet}. These sentences are appended together to form the final scene description. 

\subsection{Results}
Additional text-conditioned scenes are shown in Fig.~\ref{fig:text_scene}. Since the room shape is not given as input, objects are placed within a room-shape prior learnt from the ground truth scenes.

\begin{figure*}
\centering\begin{tabular}{@{}c@{ }c@{ }c@{ }c@{ }}
\captionsetup{justification=centering}
\subfloat[The room has an ottoman and a wardrobe cabinet.]{\includegraphics[width=.3\linewidth]{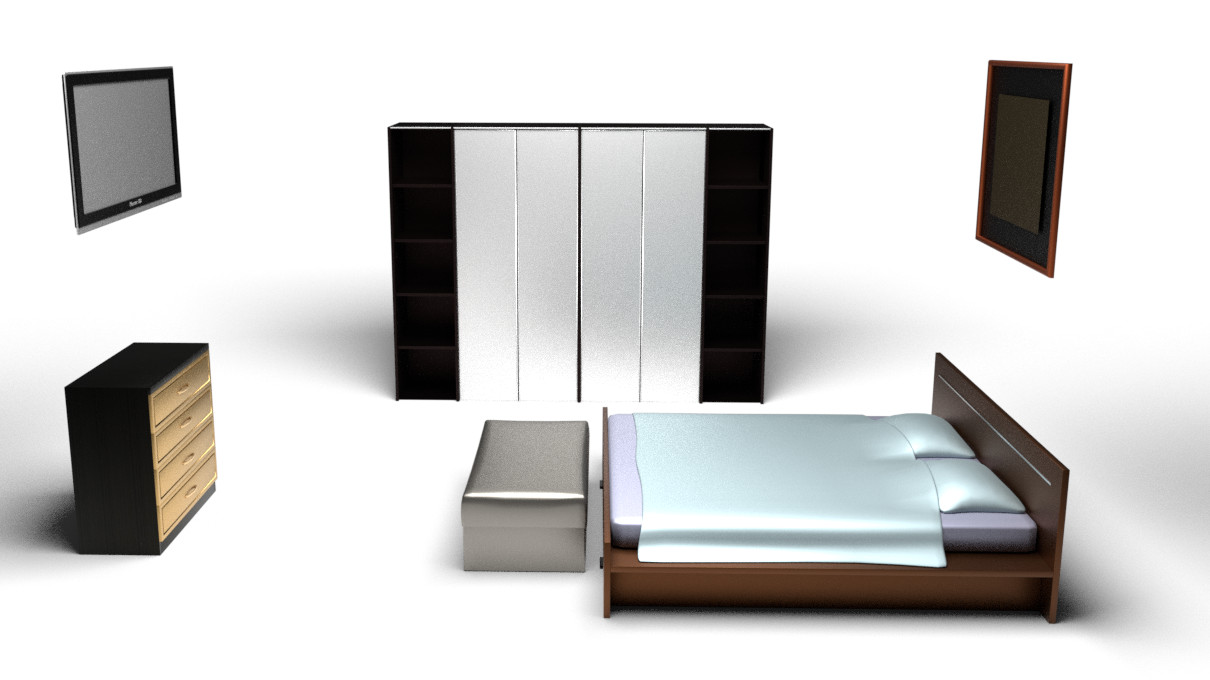}} &

\captionsetup{justification=centering}
\subfloat[In the room there are a sofa chair, a wardrobe cabinet and a double bed. The double bed is behind the sofa chair.]{\includegraphics[width=.3\linewidth]{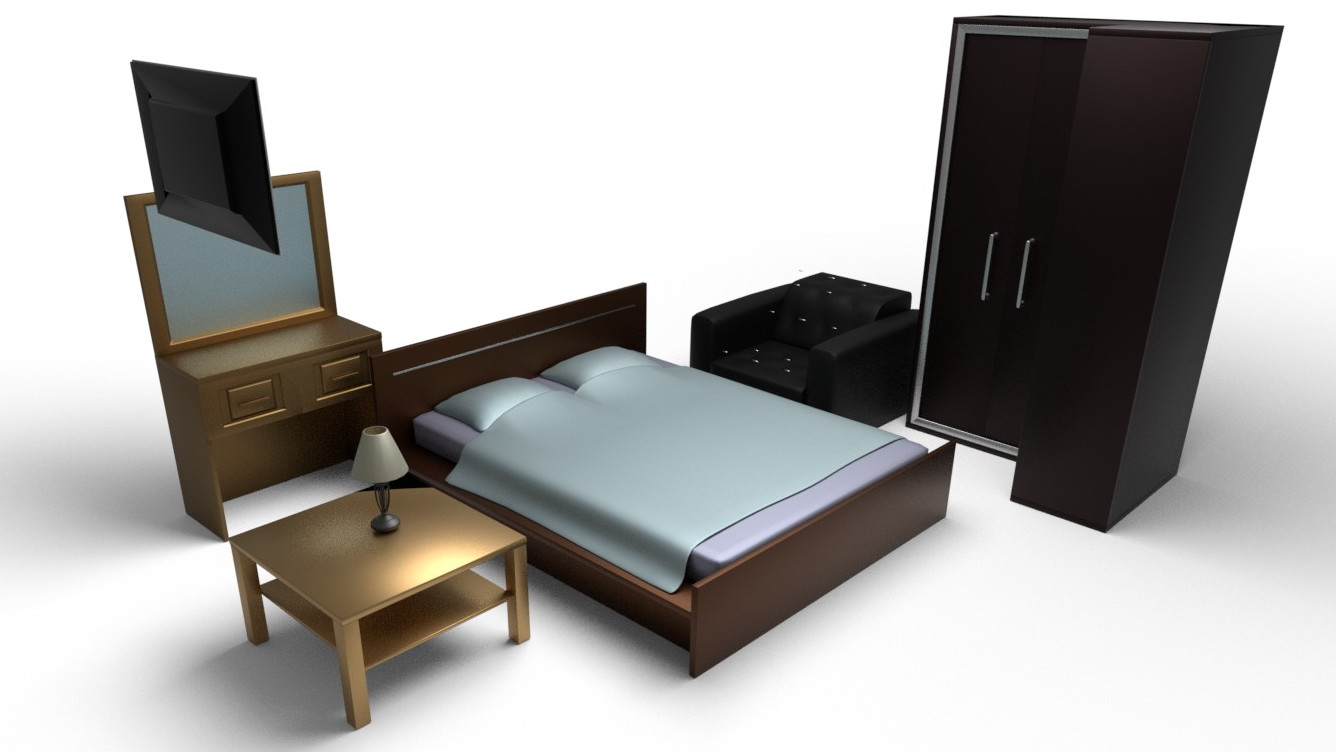}}
 &

\captionsetup{justification=centering}
\subfloat[The room has a shelving and two wardrobe cabinets.]{\includegraphics[width=.3\linewidth]{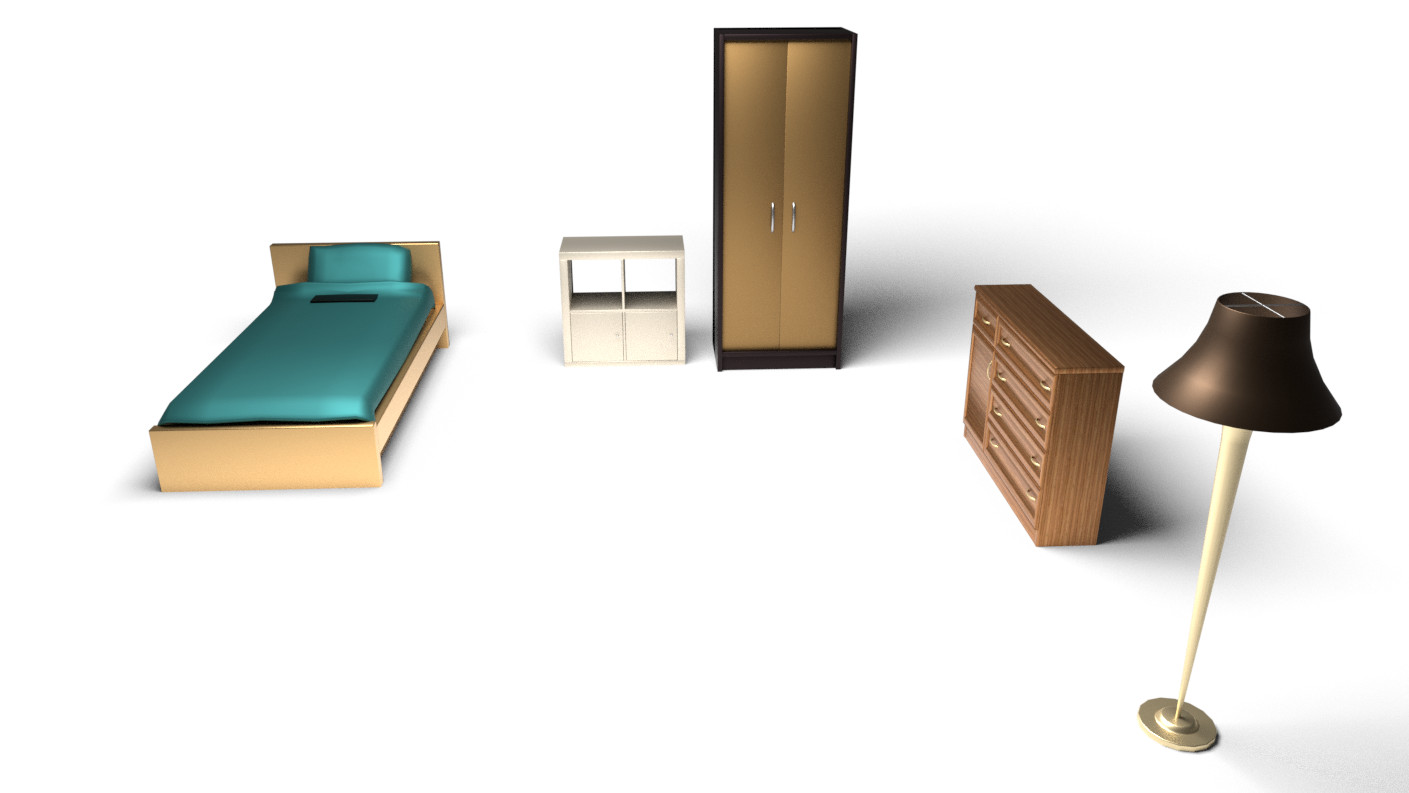}} 

\\
\captionsetup{justification=centering}
\subfloat[This room has a wardrobe cabinet and a double bed. There is a desk to the left of the double bed.
]{\includegraphics[width=.3\linewidth]{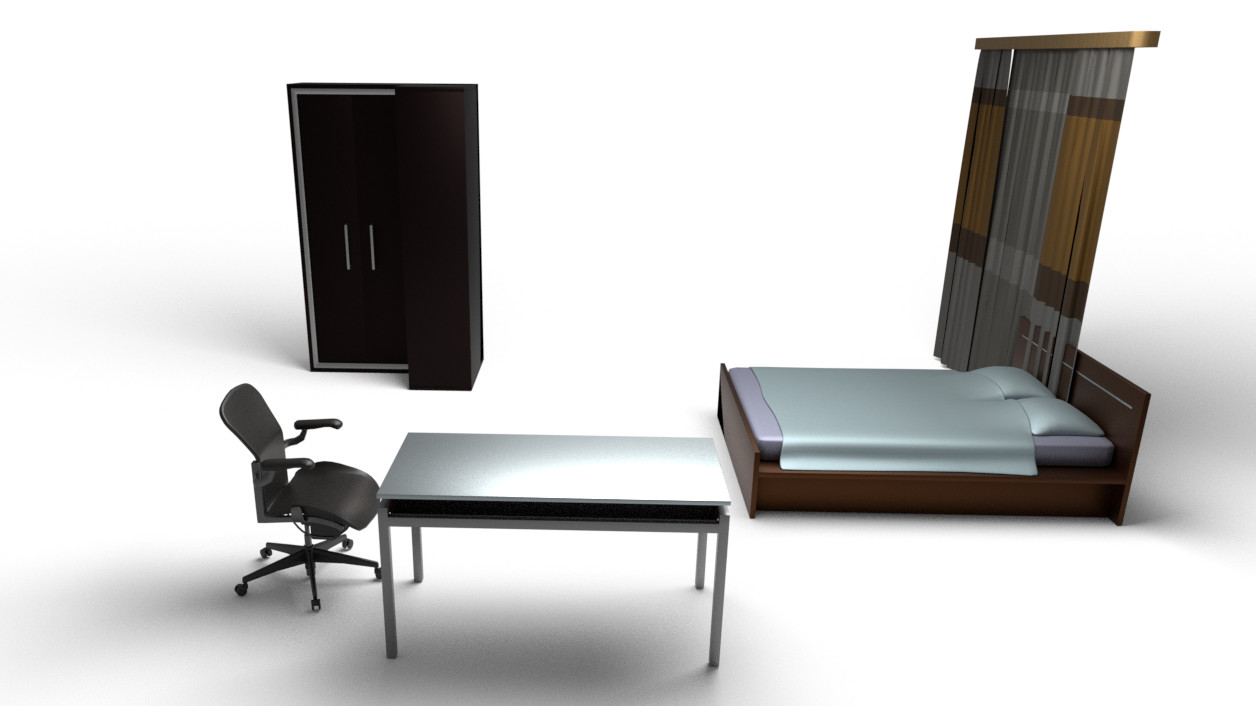}}

 &

\captionsetup{justification=centering}
\subfloat[The room has two stands and a double bed. The double bed is to the right of the second stand. There is a dresser to the right of the double bed.
]{\includegraphics[width=.3\linewidth]{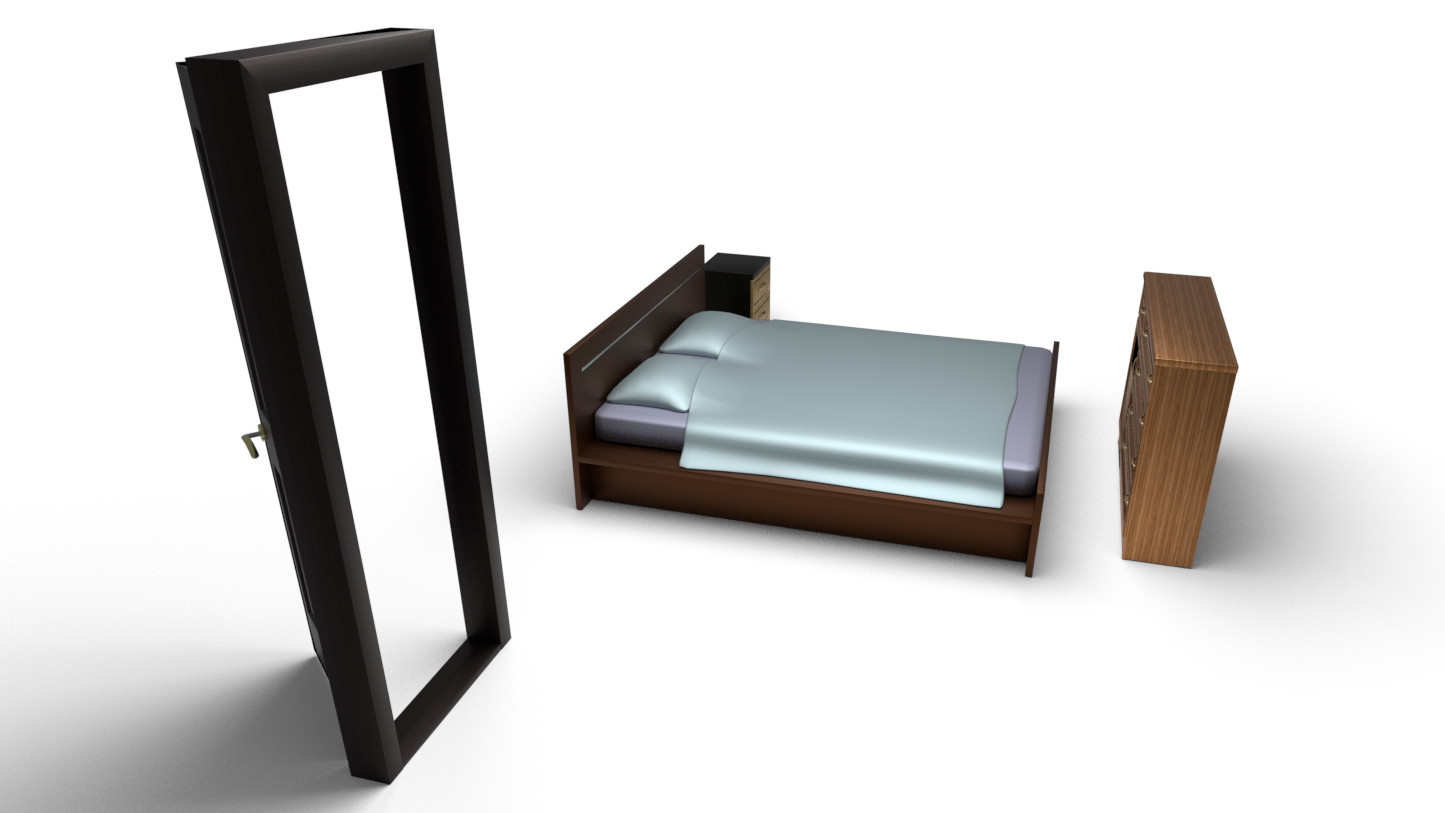}} &

\captionsetup{justification=centering}
\subfloat[In the room we see a wardrobe cabinet, a double bed and a television. There is a dresser to the left of the television.]{\includegraphics[width=.3\linewidth]{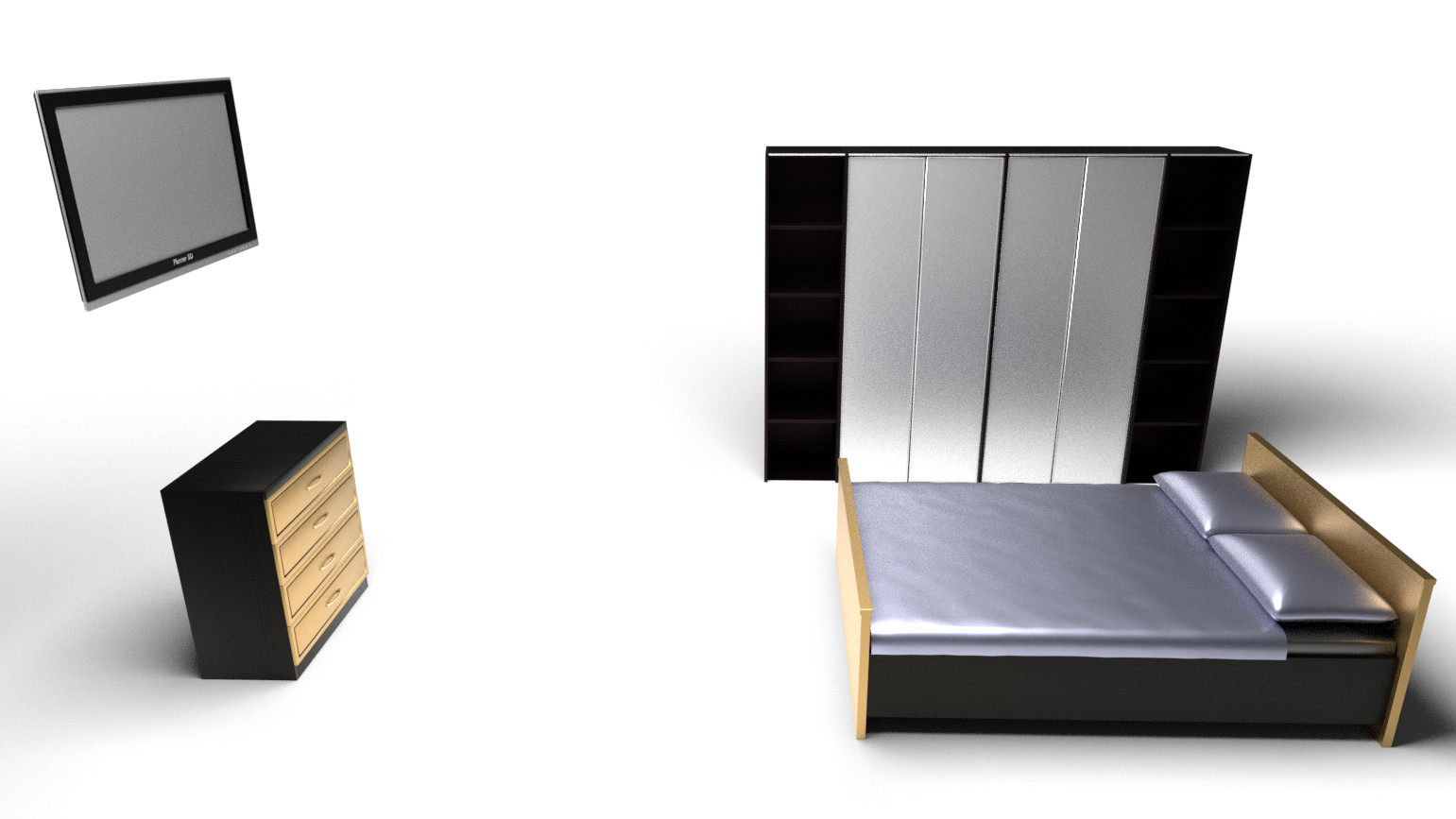}}

\end{tabular}
\caption{Text-conditioned bedroom scenes shown with the corresponding text inputs.  Important relations such as table-chair, bed-television and sofa-television are captured by the model.}

\label{fig:text_scene}
\end{figure*}

\subsection{Survey Interface}
\begin{figure*}
\centering
\includegraphics[width=0.8\textwidth]{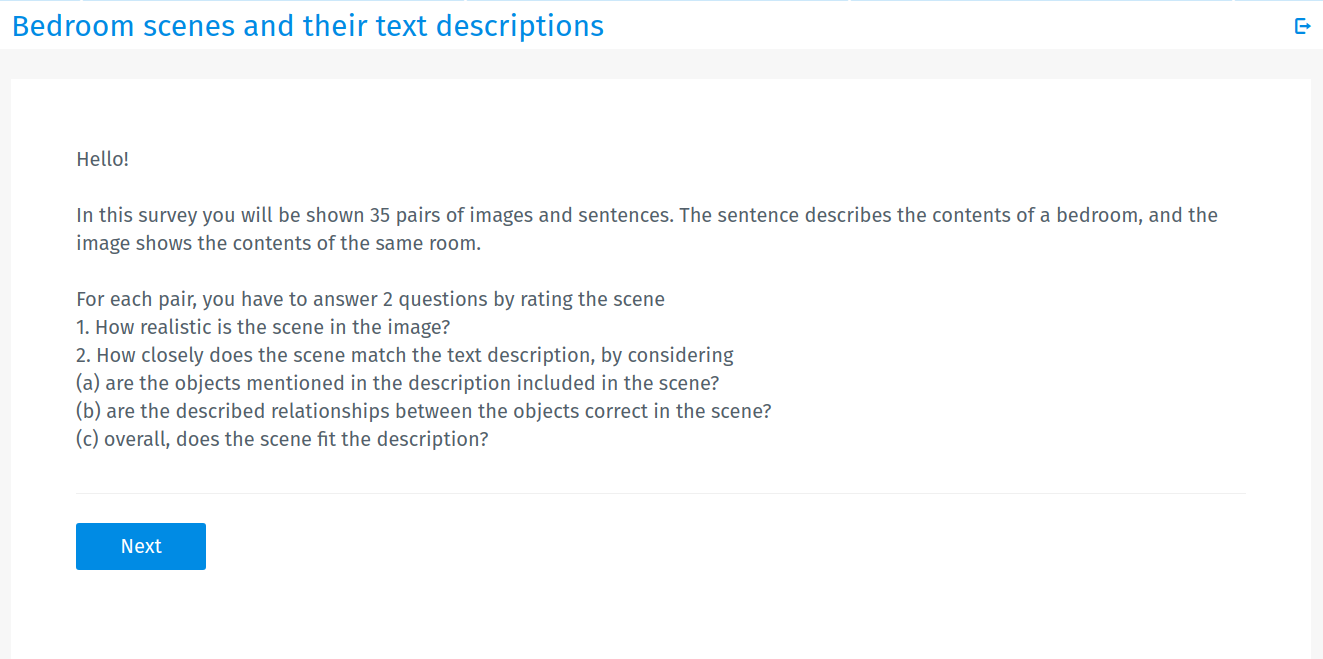}
\caption{Introduction to the perceptual study of text-conditional samples}
\label{fig:textsurvey1}
\end{figure*}

\begin{figure*}
\centering
\includegraphics[width=0.8\textwidth]{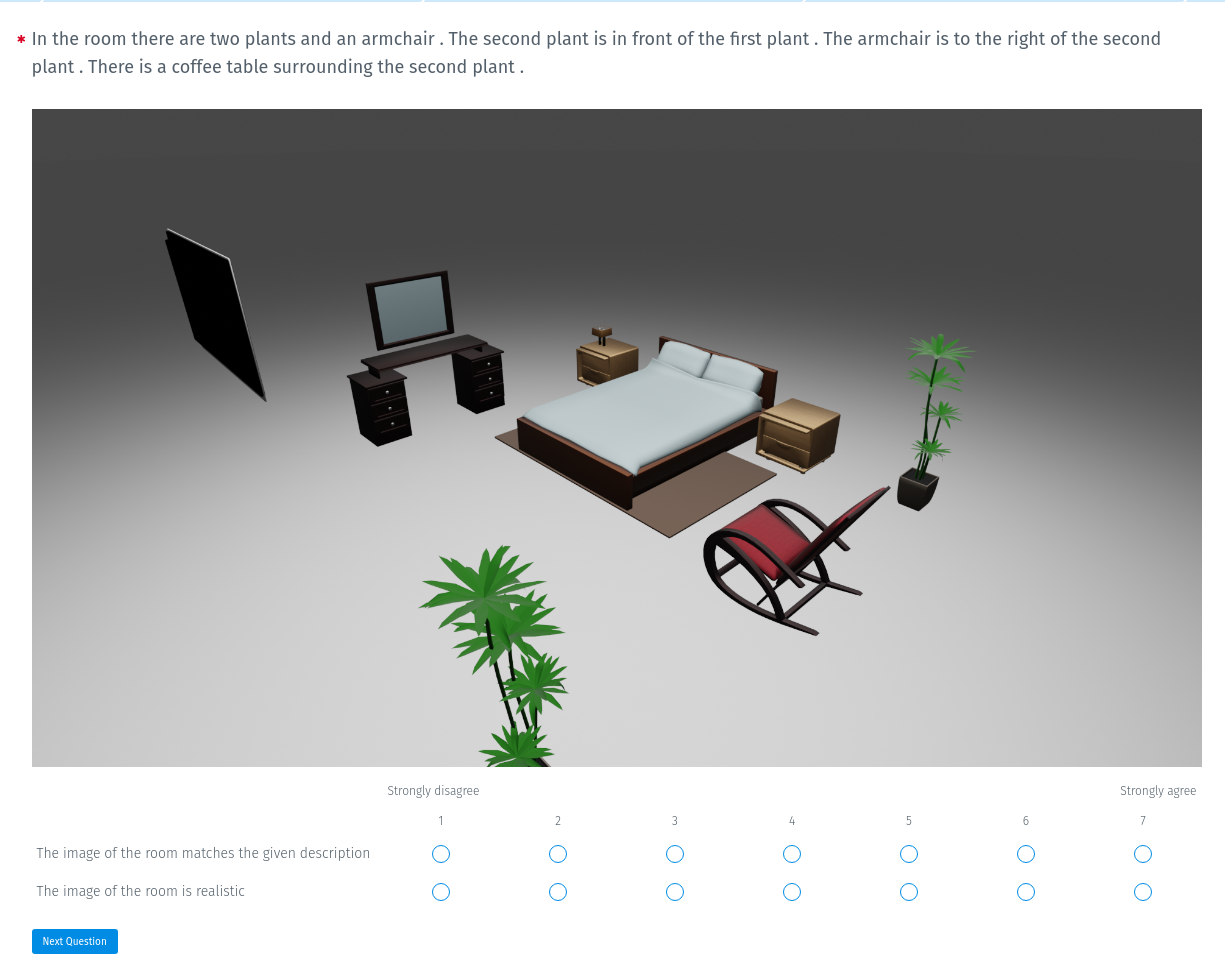}
\caption{A representative question from the perceptual study on text-conditional samples}
\label{fig:textsurvey2}
\end{figure*}

The web interface used to conduct the survey of generated scene quality is shown in Fig.~\ref{fig:textsurvey1} and Fig.~\ref{fig:textsurvey2}. The introduction explains the task to the user, and each scene has 2 questions, with answers in the range of 1--7.

\end{document}